\documentclass[runningheads]{llncs}

\usepackage{eccv}

\usepackage{eccvabbrv}

\usepackage{graphicx}
\usepackage{booktabs}

\usepackage[accsupp]{axessibility}  
\usepackage{wrapfig}

\usepackage[pagebackref,breaklinks,colorlinks,citecolor=eccvblue]{hyperref}

\usepackage{orcidlink}
\usepackage{arydshln}
\usepackage{multirow}
\usepackage{xcolor}
\usepackage{algorithmic}
\usepackage[ruled,vlined]{algorithm2e}
\SetAlFnt{\scriptsize}
\definecolor{commentcolor}{RGB}{110,154,155}   
\newcommand{\PyComment}[1]{\ttfamily\textcolor{commentcolor}{\# #1}}  
\newcommand{\PyCode}[1]{\ttfamily\textcolor{black}{#1}} 

\begin{document}

\title{Decoupling Common and Unique Representations for Multimodal Self-supervised Learning} 

\titlerunning{DeCUR for multimodal SSL}

\author{Yi Wang\inst{1,2}\orcidlink{0000-0002-3096-6610} \and
Conrad M Albrecht\inst{2}\orcidlink{0009-0009-2422-7289} \and
Nassim Ait Ali Braham\inst{1,2}\orcidlink{0009-0001-3346-3373} \and
Chenying Liu\inst{1}\orcidlink{0000-0001-9172-3586} \and
Zhitong Xiong\inst{1}\orcidlink{0000-0002-3953-585X} \and
Xiao Xiang Zhu\inst{1,3}\orcidlink{0000-0001-5530-3613}
}

\authorrunning{Y.~Wang et al.}

\institute{Data Science in Earth Observation, Technical University of Munich, Germany
\email{\{yi4.wang, chenying.liu, zhitong.xiong, xiaoxiang.zhu\}@tum.de}
\and
Remote Sensing Technology Institute, German Aerospace Center, Germany \and
Munich Center for Machine Learning, Germany\\
\email{\{conrad.albrecht,nassim.aitalibraham\}@dlr.de}
}

\maketitle

\begin{abstract}

The increasing availability of multi-sensor data sparks wide interest in multimodal self-supervised learning. However, most existing approaches learn only common representations across modalities while ignoring intra-modal training and modality-unique representations. We propose \textbf{De}coupling \textbf{C}ommon and \textbf{U}nique \textbf{R}epresentations (DeCUR), a simple yet effective method for multimodal self-supervised learning. By distinguishing inter- and intra-modal embeddings through multimodal redundancy reduction, DeCUR can integrate complementary information across different modalities. 
We evaluate DeCUR in three common multimodal scenarios (radar-optical, RGB-elevation, and RGB-depth), and demonstrate its consistent improvement regardless of architectures and for both multimodal and modality-missing settings. With thorough experiments and comprehensive analysis, we hope this work can provide valuable insights and raise more interest in researching the hidden relationships of multimodal representations\footnote{\url{https://github.com/zhu-xlab/DeCUR}}. 

\end{abstract}

\section{Introduction}

Self-supervised learning has achieved breakthroughs in machine learning \cite{ericsson2022self} and many other communities \cite{krishnan2022self, wang2022self}. Driven by the success of single-modal representation learning, as well as the great potential that large-scale multi-sensor data bears, multimodal self-supervised learning is gaining increasing attention \cite{radford2021learning,akbari2021vatt,xiong2024neural,wei2022mvp,mu2022slip,xiong2024one}. A common strategy for existing works is to align different modalities as augmented views and conduct cross-modal contrastive learning, pulling together features of different modalities for the same scene and pushing away those for the different scenes.

\begin{figure}[t]
\centering
\includegraphics[width=0.75\linewidth]{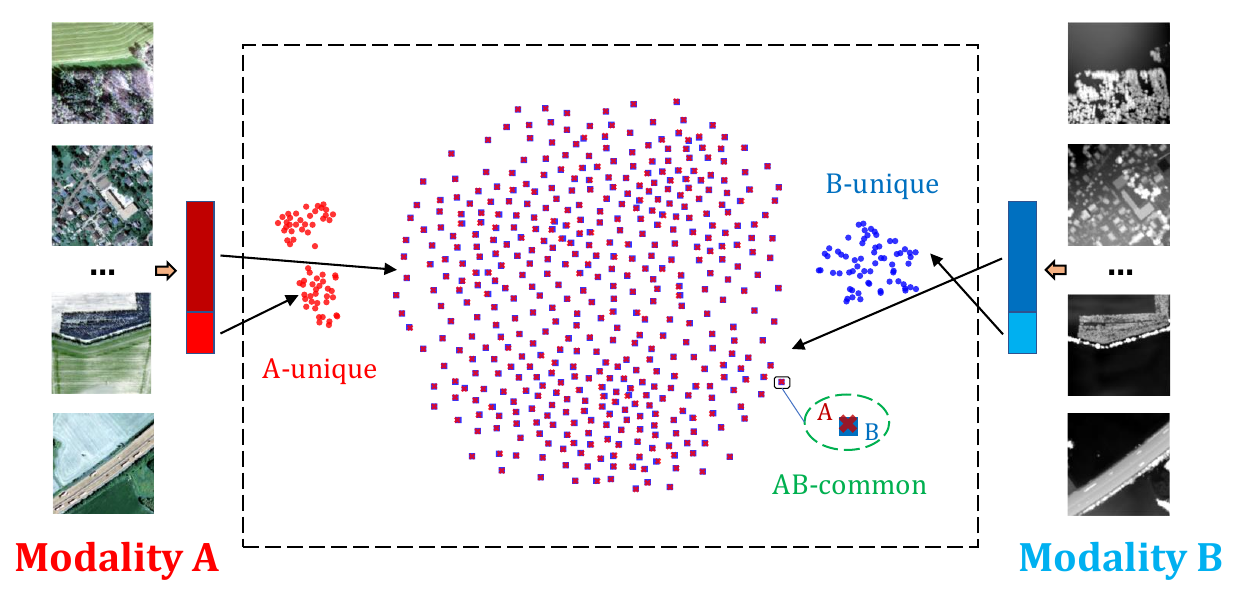}
\caption{Decoupled common and unique representations across two modalities visualized by t-SNE \cite{van2008visualizing}. \textbf{Each embedding dimension is one data point.} Red and blue circles indicate unique features from modalities A and B; red cross and blue square indicate common features from A and B. The figure shows that common and unique features from different modalities are well separated in the embedding space, and the common features between modalities are well overlapped. Best view in color \& zoomed in.}
\label{fig:decur-p1}
\end{figure}

While aligning different modalities in a common latent space has shown success in various multimodal scenarios \cite{scheibenreif2022self,radford2021learning,wang2021multimodal,Girdhar_2023_CVPR}, the fact that one modality may hold unique information that can not be extracted from other modalities tend to be overlooked. As a result, such modality-unique information is suppressed during training. This forces the model to put potentially orthogonal representations into common feature embeddings, limiting the model's capacity to understand different modalities in detail. To tackle this issue, we introduce the idea of Decoupling Common and Unique Representations (DeCUR) for multi-modal self-supervised learning.

Specifically, we separate the feature embedding dimensions into cross-modal common ones and modality-unique ones. During training, we calculate the normalized cross-correlation matrix of the common dimensions and the unique dimensions between two modalities, and drive the matrix of the common dimensions to the identity, while the matrix of the unique dimensions to zero. As a result, common embeddings are aligned across modalities, while modality-unique embeddings are pushed away. In practice, this can be seen as a natural extension of Barlow Twins \cite{zbontar2021barlow}, one self-supervised learning technique that conducts redundancy reduction on the dimension level.


However, simply pushing away unique dimensions will lead to a collapse that these dimensions do not learn any useful information. Therefore, apart from cross-modal learning, we include also intra-modal learning which utilizes all embedding dimensions and drives the cross-correlation matrix between two augmented views of the same modality to the identity. This intra-modal component not only avoids the collapse by letting the unique dimensions learn meaningful representations within one modality, but also enhances cross-modal learning with stronger intra-modal knowledge. \cref{fig:decur-p1} provides a t-SNE \cite{van2008visualizing} visualization of the latent space of the learned representations, where common and unique embeddings of each modality, as well as modality-unique embeddings between modalities, are well separated.

In addition, previous works have shown that different modalities may embed important information in different regions of the feature maps \cite{xiong2020msn,zhang2022cmx}. Inspired by these findings, we introduce a modern deformable attention module \cite{xia2022vision,xia2023dat++} in our ConvNet backbones to help the model focus on modality-specific important areas during training. Such deformable attention selects the positions of key and value pairs in self-attention in a data-dependent way, which is both more efficient compared to previous attention modules and well-fitted for our modality-enhancing purpose. In summary, our main contributions are listed as follows:
\begin{itemize}
   \item We propose DeCUR, a simple yet effective multimodal self-supervised learning method, which decouples common and unique representations across different modalities and enhances both intra- and inter-modal learning. For ConvNet backbones, we adopt a simple adaptation of deformable attention for modality-informative feature learning.  
   \item We evaluate DeCUR with rich experiments and comprehensive analysis covering three important multi-modal scenarios, demonstrating its effectiveness in both multi-modal and modality-missing settings.
\end{itemize}

\section{Related work}

\subsubsection{Self-supervised learning} 
Self-supervised learning of single modality has been widely studied, which can be categorized into three main types: generative methods such as Autoencoder \cite{vincent2010stacked} and Masked Autoencoder \cite{he2022masked}, predictive methods such as predicting rotation angles \cite{gidaris2018unsupervised}, and contrastive methods that train joint embedding architectures with or without negative samples. 
Contrastive methods can be further categorized into four strategies of self-supervision: 1) contrastive learning with negative samples such as CPC \cite{oord2018representation}, SimCLR \cite{chen2020simple} and MoCo \cite{he2020momentum}; 2) clustering feature embeddings such as SwAV \cite{caron2020unsupervised}; 3) knowledge distillation such as BYOL \cite{grill2020bootstrap}, SimSiam \cite{chen2021exploring} and DINO \cite{caron2021emerging}; 4) redundancy reduction such as Barlow Twins \cite{zbontar2021barlow} and VICReg \cite{bardes2021vicreg}. 
The second and third categories usually require common encoders, thus not easily adaptable for modality-specific encoders in multimodal scenarios. 
While most existing multimodal works are closely related to the first strategy, DeCUR belongs to redundancy reduction as a natural extension of Barlow Twins that does not require numerous negative samples. Specifically, DeCUR's decoupling strategy can be perfectly integrated into a simple correlation-matrix-based loss design in Barlow Twins. 

\subsubsection{Multimodal self-supervised learning} 
The idea of contrastive self-supervised learning can be easily transferred to multimodal scenarios, as different modalities are naturally the augmented views.
Currently, contrastive learning with negative samples has been mostly developed: CLIP \cite{radford2021learning} for language-image, VATT \cite{akbari2021vatt} for video-audio-text, CROMA \cite{fuller2024croma} for radar-optical, and ImageBind \cite{Girdhar_2023_CVPR} for a joint embedding of six different modalities.
Different from these methods, we propose to explore the potential of negative-free methods by extending the redundancy reduction loss of Barlow Twins. We also take one step further to decouple common and unique information from different modalities. Meanwhile, we share an insight with Yang \etal \cite{yang2022vision} and Wang \etal \cite{wang2022self1} that intra-modal representations are important complements to cross-modal representations.

\subsubsection{Modality decoupling} 
While not well explored in self-supervised research, modality decoupling has been proven beneficial in multimodal supervised learning. Xiong \etal \cite{xiong2020msn,xiong2021ask} studied multimodal fusion from network architecture, proposing modality separation networks for RGB-D scene recognition. Peng \etal \cite{peng2022balanced} investigated modality dominance from the angle of optimization flow, proposing on-the-fly gradient modulation to balance and control the optimization of each modality in audio-visual learning. Zhou \etal \cite{zhou2023feature} observed feature redundancy for different supervision tasks, proposing to decompose task-specific and task-shared features for multitask learning in recommendation system. FactorCL \cite{liang2024factorized} studies the decoupling concept in self-supervision, factorizing task-relevant information into shared and unique representations with modality-specific augmentations. Different from the above, we directly perform modality decoupling on the embeddings by separating common and unique dimensions.

\section{Methodology}

\cref{fig:decur} presents the general structure of DeCUR. As a multimodal extension of Barlow Twins, DeCUR performs self-supervised learning by redundancy reduction in the joint embedding space of augmented views from both intra-/cross-modal perspectives. Here, our main contribution lies in a simple loss design to decouple meaningful modality-unique representations across modalities. 

\subsection{Decoupling common and unique representations}

\begin{figure}[hbtp]
\centering
\includegraphics[width=1\linewidth]{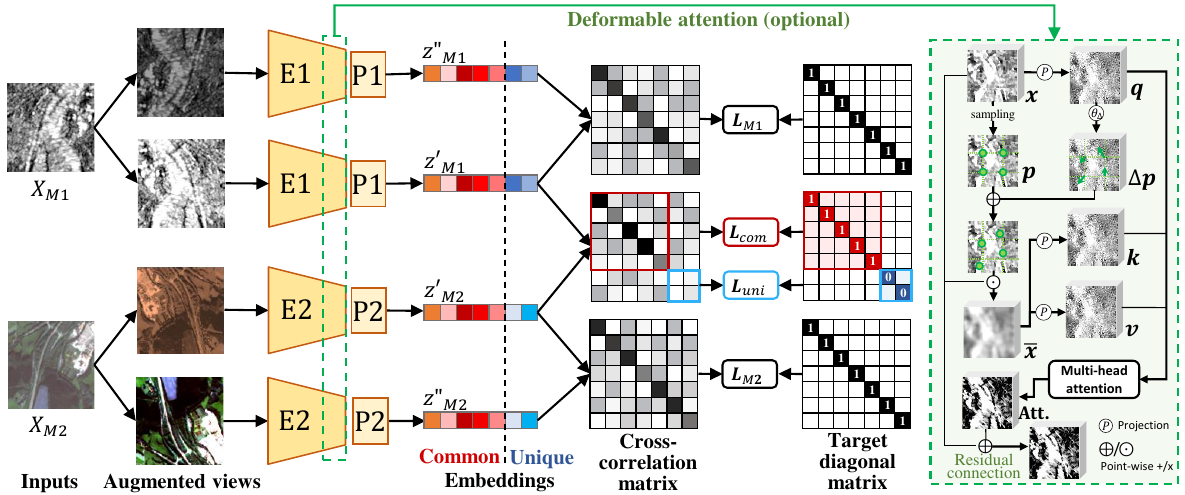}
\caption{
The structure of DeCUR. \textit{$M1$} and \textit{$M2$} represent two modalities. Two augmented views from each modality are fed to modality-specific encoders ($E1$, $E2$) and projectors ($P1$, $P2$) to get the embeddings $Z$. 
For cross-modal embeddings, the dimensions are separated into \textcolor{red}{common} and \textcolor{blue}{unique} parts. The correlation matrix of the common dimensions is optimized to be close to the identity, while that of the unique ones to zero. For intra-modal embeddings, both common and unique dimensions are used to calculate the correlation matrix which is optimized to be close to the identity. 
DeCUR optionally adds deformable attention (the green shadowed region on the right side) in the last layers of ConvNet encoders to boost modality-informative learning.
}
\label{fig:decur}
\end{figure}

As shown in \cref{fig:decur}, we feed two batches of augmented views of the inputs from each modality into the modality-specific encoders and projectors, producing corresponding embeddings ${Z_{M1}}'$ and ${Z_{M1}}''$ for $X_{M1}$, and ${Z_{M2}}'$ and ${Z_{M2}}''$ for $X_{M2}$, respectively. Batch normalization is applied on embeddings such that they are mean-centered along the batch dimension. These embeddings are then used to calculate cross-correlation matrices across/within modalities for optimization.

\subsubsection{Cross-correlation matrix} 
Given two embedding vectors $Z^{A}, Z^{B} \in \mathbb{R}^{K}$, the cross-correlation matrices $\mathcal{C}$ between them is formulated as \cite{zbontar2021barlow}:
\begin{equation}
\mathcal{C}_{i j} = \frac{\sum_b z_{b, i}^A z_{b, j}^B}{\sqrt{\sum_b\left(z_{b, i}^A\right)^2} \sqrt{\sum_b\left(z_{b, j}^B\right)^2}}\quad
\end{equation}
where $b$ indexes batch samples, and $i$, $j$ index the dimension of the embedding vectors. $\mathcal{C} \in \mathbb{R}^{K \times K}$ is a square matrix with values ranging from -1 to 1.

\subsubsection{Cross-modal representation decoupling} 
In the cross-modal case, the correlation matrix $\mathcal{C}$ is calculated between two embeddings from different modalities, such as ${Z_{M1}}'$ and ${Z_{M2}}'$ in \cref{fig:decur}.
While most multimodal self-supervised learning algorithms consider only their common representations, we explicitly consider the existence of modality-unique representations and decouple them during training. Specifically, we separate the total embedding dimension $K$ into $K_c$ and $K_u$ with $K_c+K_u=K$ to store common and unique representations, respectively. The common representations should be identical across modalities (red parts in \cref{fig:decur}), while the modality-specific unique representations should be decorrelated (blue parts in \cref{fig:decur}).

Specifically, on the one hand, to learn cross-modal common information, a sub-matrix $\mathcal{C_\text{c}} \in \mathbb{R}^{K_c \times K_c}$ is generated from only the common dimensions of ${Z_{M1}}'$ and ${Z_{M2}}'$. The redundancy reduction loss for the cross-modal common representations reads:

\begin{equation}
\mathcal{L}_{com} = \sum_i\left(1-\mathcal{C_\text{c}}_{i i}\right)^2+\lambda_{c} \cdot \sum_i \sum_{j \neq i} \mathcal{C_\text{c}}_{i j}^2\quad,
\end{equation}

\noindent where $\lambda_{c}$ is a positive constant trading off the importance of the first invariance term (to make the common embeddings invariant to the input modalities) and the second redundancy reduction term (to decorrelate the embedding vector components and avoid model collapse).

On the other hand, to decouple modality-specific information, a sub-matrix $\mathcal{C_\text{u}} \in \mathbb{R}^{K_u \times K_u}$ is generated from the unique dimensions of ${Z_{M1}}'$ and ${Z_{M2}}'$. The redundancy reduction loss for the modality-unique representations reads:

\begin{equation}
\mathcal{L}_{uni} = \sum_i\mathcal{C_\text{u}}_{i i}^2+\lambda_{u} \cdot \sum_i \sum_{j \neq i} \mathcal{C_\text{u}}_{i j}^2 \quad,
\end{equation}

\noindent where $\lambda_{u}$ is a positive constant trading off the importance of the first decorrelation term (to decorrelate different modalities) and the second redundancy reduction term (to decorrelate the embedding vector components). However, pure decoupling doesn't ensure the meaningfulness of the unique dimensions, i.e., they could collapse into random decorrelated values. To tackle this issue, we further introduce intra-modal representation enhancement that covers both common and unique dimensions within each modality.

\subsubsection{Intra-modal representation enhancing} 
To avoid the collapse of the decoupled unique dimensions in the cross-modal training, as well as to boost intra-modal representations, we introduce intra-modal training that covers all the embedding dimensions.
For each modality, a cross-correlation matrix $\mathcal{C_\text{M1}}$ (or $\mathcal{C_\text{M2}}$) is generated from the full dimensions of the embedding vectors ${Z_{M1}}'$ and ${Z_{M1}}''$ (or ${Z_{M2}}'$ and ${Z_{M2}}''$). The redundancy reduction losses for the intra-modal representations read:

\begin{equation}
\mathcal{L}_{M1} = \sum_i\left(1-\mathcal{C_\text{M1}}_{i i}\right)^2+\lambda_{M1} \cdot \sum_i \sum_{j \neq i} \mathcal{C_\text{M1}}_{i j}^2 \quad,
\end{equation}
\begin{equation}
\mathcal{L}_{M2} = \sum_i\left(1-\mathcal{C_\text{M2}}_{i i}\right)^2+\lambda_{M2} \cdot \sum_i \sum_{j \neq i} \mathcal{C_\text{M2}}_{i j}^2\quad,
\end{equation}

\noindent where $\lambda_{M1}$ and $\lambda_{M2}$ are positive constants trading off the importance of the invariance term and the redundancy reduction term.

Combining the cross-modal common and unique losses, and the intra-modal losses, the overall training objective of DeCUR reads:

\begin{equation}
\mathcal{L} = \mathcal{L}_{com} + \mathcal{L}_{uni} + \mathcal{L}_{M1} + \mathcal{L}_{M2}\quad.
\end{equation}


\subsection{Deformable attention for modality-informative features}

Apart from the DeCUR loss design, we adopt deformable attention to help ConvNet models focus on modality-informative regions. The deformable attention module was proposed in DAT \cite{xia2022vision} and DAT++ \cite{xia2023dat++} to efficiently model the relations among feature tokens under the guidance of the important regions in the feature maps. The readers are referred to the original papers for technical details, while a brief simplified recap is as follows. Given an input feature map $x \in \mathbb{R}^{H \times W \times C}$, a downsampled grid of points $p \in \mathbb{R}^{H_G \times W_G \times 2}$ is generated as references, where $H_G = H/r$ with $r$ being the downscaling ratio. In parallel, the feature map $x$ is projected to the query tokens $q \in \mathbb{R}^{H \times W \times C}$, where $q=xW_q$. The query tokens $q$ are fed into a lightweight sub-network 
${\theta}_{\text{offset}}$ to generate the offsets $\Delta{p} \in \mathbb{R}^{H_G \times W_G \times 2}$ in order to get final deformed points with $p+\Delta{p}$. Then the features are sampled from $x$ at the locations of deformed points and interpolated to a feature map $\bar{x} \in \mathbb{R}^{H_G \times W_G \times C}$. This sampled feature map $\bar{x}$ is projected to keys $k$ and values $v$, where $k=\bar{x}W_k$ and $v=\bar{x}W_v$. Softmax attention is then calculated on flattened queries $q$ and keys $k$ and multiplied with values $v$. The final output is reshaped back to the same size as the input feature map $x$. 

We adopt the deformable attention module in the last two stages of the encoder to learn regional focus while keeping efficiency. A residual connection from the input feature map to the output of the deformable attention module is added to restrict unexpected influences of the attention module, such as biasing the pretraining towards the pretext task by selecting unexpected deformable points. This is especially helpful in early training, when the model needs to first capture general information. With the training going on, the model then gradually learns detailed modality-specific representations (c.f. \cref{sec:abexp}). 

\section{Implementation details}
\label{sec:implementation}
\subsubsection{Pretraining datasets} 
We pretrain DeCUR in three multimodal scenarios: SAR-optical, RGB-DEM and RGB-depth. For SAR-optical, we use the SSL4EO-S12 dataset \cite{wang2022ssl4eo} which consists of 251k multi-modal image pairs from multiple seasons with size 264x264. One random season is selected for each image and modality. For RGB-DEM, we conduct pretraining on the training set of GeoNRW dataset \cite{s5xq-b822-20} which includes aerial RGB images, digital elevation models (DEM) and segmentation maps from the German state North Rhine-Westphalia. We crop the raw 6,942 training scenes to 111k patches with size 250x250. For RGB-depth, we use SUN-RGBD dataset \cite{song2015sun} which consists of 10,335 RGBD pairs with various image sizes. We preprocess the depth images to HHA format \cite{gupta2014learning} following \cite{zhang2022cmx}. Common data augmentations \cite{grill2020bootstrap} are selected and used based on the feasibility in each specific modality (c.f. Appendix).


\subsubsection{Model architecture}
As a multimodal extension of Barlow Twins \cite{zbontar2021barlow}, each branch holds a separate backbone and a 3-layer MLP projector (each with output dimension 8192). DeCUR is trained on embedding representations after the projector, whose dimensions are separated into common and unique. We do a light grid search to get the best corresponding ratio. For SAR-optical, the percentage of common dimensions is 87.5\%; for RGB-DEM and RGB-depth it is 75\%. The backbones are transferred to downstream tasks. We use ResNet-50 \cite{he2016deep} for all scenarios, with additional MiT-B2/B5 from SegFormer \cite{xie2021segformer} for RGB-Depth. We adopt deformable attention to the last two stages of the ResNet-50 backbone. We do not adopt deformable attention to MiTs as attention is already integrated.

\subsubsection{Optimization}
We follow the optimization protocol of Barlow Twins \cite{zbontar2021barlow} and BYOL \cite{grill2020bootstrap}, with default epochs 100 and a batch size of 256 for SAR-optical and RGB-DEM (epochs 200 and batch size 128 for RGB-depth). The trade-off parameters $\lambda$ of the loss terms are set to 0.0051. Training is distributed across 4 NVIDIA A100 GPUs and takes about 35 hours on SSL4EO-S12, 6 hours on GeoNRW, and 6 hours on SUN-RGBD.

\section{Experimental results}

We evaluate DeCUR by transferring to three common multimodal tasks: SAR-optical scene classification, RGB-DEM semantic segmentation, and RGB-depth semantic segmentation. We follow common evaluation protocols of frozen encoder and fine-tuning. We report results for full- and limited-label settings, and both multimodal and modality-missing (single-modal) settings.

\subsection{SAR-optical scene classification}
We pretrain SAR-optical encoders on SSL4EO-S12 \cite{wang2022ssl4eo} and transfer them to BigEarthNet-MM \cite{sumbul2021bigearthnet}, a multimodal multi-label scene classification dataset with 19 classes.
Simple late fusion is used for multimodal transfer learning by concatenating the encoded features from both modalities, followed by one classification layer. Mean average precision (mAP, micro) serves as the evaluation metric.

\begin{table}[tb]
\caption{SAR-optical transfer results (mAP) on BigEarthNet-MM (left: multimodal; middle: SAR-only), and optical-only results on BigEarthNet-S2 compared with SOTA Earth observation foundation models (right). We report both "linear-probing/fine-tuning" scores. \textit{Rand. Init.} represents random initialization, \textit{-cross} represents cross-modal, \textit{BT-SAR} represents Barlow Twins with SAR-only. The same denotations are used in the following. Best scores among self-supervised methods are marked in \textbf{bold}. *: SkySense uses a mixed backbone combining ViT and Swin Transformer.  
}
\label{tab:BE-mm}
\centering
\scalebox{0.68}{
\begin{tabular}{lcc}
\toprule
SAR-optical & 1\% labels & 100\% labels \\ \hline
Rand. Init.                     & 58.7       & 70.1         \\
Supervised                      & 77.0       & 88.9         \\
\hdashline
SimCLR-cross          & 77.4/78.7  & 82.8/89.6    \\
CLIP                            & 77.4/78.7  & 82.8/89.6    \\
Barlow Twins                    & 78.7/80.3  & 83.2/89.5    \\
VICReg                          & 74.5/79.0  & 81.9/89.5    \\
DeCUR (ours)                    & \textbf{79.8}/\textbf{81.5}  & \textbf{86.2}/\textbf{89.8}    \\ \bottomrule
\end{tabular}
}
\scalebox{0.68}{
\begin{tabular}{lcc}
\toprule
SAR-only & 1\% labels & 100\% labels    \\ \hline
Rand. Init.          & 50.0           & 54.2           \\
Supervised           & 67.5           & 81.9           \\ \hdashline
SimCLR-cross         & 68.1/70.4      & 71.7/83.7  \\
{CLIP}  & {68.0/70.2} & {71.7/83.4} \\
Barlow Twins         & 72.3/73.7          & {77.8/83.6} \\
VICReg       & 69.3/71.9             & {74.1/83.6} \\
BT-SAR      & 71.2/73.3          & 77.5/81.6           \\
DeCUR (ours)         & \textbf{74.4}/\textbf{76.0} & \textbf{79.5}/\textbf{84.0}  \\ \bottomrule
\end{tabular}
}
\scalebox{0.65}{
\begin{tabular}{llc}
\toprule
Optical-only      & Backbone      & 10\% labels   \\ \hline
SeCo \cite{manas2021seasonal}         & RN50    & 78.6/82.6        \\
SSL4EO \cite{wang2023ssl4eo}       & RN50    & 82.1/86.2        \\
FG-MAE \cite{wang2023feature}       & ViT-S   & 78.1/85.2        \\
\hdashline
GFM \cite{mendieta2023towards}    & Swin-B & -/86.3           \\
SpectralGPT \cite{hong2023spectralgpt}  & ViT-B  & -/87.5           \\
SatMAE \cite{cong2022satmae}       & ViT-L  & 80.3/86.2        \\
CROMA \cite{fuller2024croma}        & ViT-L  & \textbf{85/88.3} \\
SkySense \cite{guo2023skysense}     & ViT-L* & \textbf{-/88.7}  \\
\hdashline
DeCUR (ours) & RN50    & 83.3/87.2        \\ \bottomrule
\end{tabular}
}
\end{table}

We report multimodal linear probing and fine-tuning results with 1\% and 100\% training labels in \cref{tab:BE-mm} (left). DeCUR outperforms existing cross-modal SimCLR-like contrastive learning by 2.4\%-3.4\% in linear probing, while achieving comparable performance on fine-tuning with full labels. Compared to BarlowTwins, we improve by 1.1\% and 1.2\% on linear evaluation and fine-tuning with 1\% labels, and 3.0\% and 0.3\% with full labels.

We report SAR-only results in \cref{tab:BE-mm} (middle), as it is an essential scenario in practice when optical images are either unavailable or heavily covered by clouds. DeCUR outperforms other methods in most scenarios by a large margin (up to 7.8\%), while achieving slightly better performance on fine-tuning with full labels. In addition, DeCUR outperforms single-modal Barlow Twins pretraining by 2.7\%-3.2\% with 1\% labels and 2.0\%-2.4\% with full labels, indicating that joint multimodal pretraining helps the model better understand individual modalities.

In addition, we compare DeCUR with state-of-the-art Earth observation foundation models on BigEarthNet-S2 with 10\% labels in \cref{tab:BE-mm} (right). DeCUR achieves better performance than existing models with comparable model sizes such as SeCo \cite{manas2021seasonal}, SSL4EO \cite{wang2022ssl4eo} and FG-MAE \cite{wang2023feature} in both linear probing and fine-tuning. Compared to SOTA large models, DeCUR is only slightly worse than CROMA \cite{fuller2024croma}, SkySense \cite{guo2023skysense} and SpectralGPT \cite{hong2023spectralgpt} with much fewer parameters, while even better than GFM \cite{mendieta2023towards} and SatMAE \cite{cong2022satmae}.

\subsection{RGB-DEM semantic segmentation}

We pretrain and evaluate RGB-DEM encoders on GeoNRW \cite{s5xq-b822-20} for semantic segmentation (10 classes). For fair and visible comparison, we use simple fully convolutional networks (FCN) \cite{long2015fully} as the segmentation model, which concatenates the last three layer feature maps from both modalities, upsamples and sums them up to generate prediction maps. 
Similar to the classification task, we report frozen-encoder and full fine-tuning results in \cref{tab:NRW-mm} with mean Intersection over Union (mIoU) serving as the evaluation metric.

\begin{table}[t]
\caption{RGB-DEM transfer learning results (mIoU) with frozen-encoder and full fine-tuning on GeoNRW (left: multimodal; right: RGB-only). 
}
\label{tab:NRW-mm}
\centering
\scalebox{0.72}{
\begin{tabular}{lcccc}
\toprule
\multirow{2}{*}{RGB-DEM} & \multicolumn{2}{c}{1\% labels} & \multicolumn{2}{c}{100\% labels} \\
                         & Frozen         & Fine-tune     & Frozen          & Fine-tune      \\ \hline
Rand. Init.              & 14.1           & 14.1          & 23.0            & 23.0           \\
Supervised               & 22.1           & 22.1          & 44.0            & 44.0           \\ \hdashline
SimCLR-cross             & 23.0           & 30.2          & 35.2            & 47.3           \\
CLIP                     & 22.8           & 28.8          & 35.0            & 46.7           \\
Barlow Twins             & 31.2           & 33.6          & 43.0            & 48.4           \\
VICReg                   & 27.4           & 32.8          & 38.0            & 45.1           \\
DeCUR (ours)             & \textbf{34.7}  & \textbf{36.6} & \textbf{44.7}   & \textbf{48.9}  \\ \bottomrule
\end{tabular}
}
\scalebox{0.72}{
\begin{tabular}{lcccc}
\toprule
\multirow{2}{*}{RGB-only} & \multicolumn{2}{c}{1\% labels} & \multicolumn{2}{c}{100\% labels} \\
                     & Frozen         & Fine-tune     & Frozen          & Fine-tune      \\ \hline
Rand. Init.          & 14.2           & 14.2          & 18.5            & 18.5           \\
Supervised           & 17.5           & 17.5          & 38.8            & 38.8           \\ \hdashline
SimCLR-cross         & 20.1           & 25.9          & 29.6            & 42.5           \\
CLIP                 & 20.0           & 25.7          & 29.4            & 42.3           \\
Barlow Twins         & 29.4           & 33.4          & 38.0            & 45.9           \\
VICReg               & 23.7           & 28.7          & 32.4            & 41.6           \\
BarlowTwins-RGB      & 28.6           & 32.6          & 36.2            & 45.7           \\
DeCUR (ours)         & \textbf{32.2}  & \textbf{35.7} & \textbf{40.8}   & \textbf{46.7}  \\ \bottomrule
\end{tabular}
}
\end{table}

We present multimodal frozen-encoder and fine-tuning results with 1\% and 100\% training labels in \cref{tab:NRW-mm} (left). Promisingly, DeCUR outperforms other methods in all scenarios by a large margin (up to 11.7\% compared to SimCLR). Notably, DeCUR works better than Barlow Twins and VICReg by at least 3.0\% when fine-tuning with 1\% labels. 
Meanwhile, we report RGB-only results in \cref{tab:NRW-mm} (right). Again DeCUR shows a significant improvement compared to other cross-modal methods in all scenarios (up to 12.1\% compared to CLIP). In addition, DeCUR outperforms single-modal Barlow Twins by 3.6\%-4.6\% with a frozen encoder, and 1.0\%-3.1\% in fine-tuning.

\subsection{RGB-depth semantic segmentation}

We pretrain RGB-depth encoders on SUN-RGBD \cite{song2015sun} and transfer them to SUN-RGBD and NYU-Depth v2 \cite{Silberman:ECCV12} datasets for semantic segmentation (37 and 40 classes, respectively). We transfer ResNet50 to simple FCN \cite{long2015fully} and MiT-B2/B5 \cite{xie2021segformer} to the recent CMX \cite{zhang2022cmx} model. We report single- and multi-modal fine-tuning results with mIoU and overall accuracy (OA) in \cref{tab:rgbd-mm}. As observed, 
DeCUR helps improve FCN over CLIP by 4.0\% mIoU and 1.3\% OA on SUN-RGBD, and 0.8\% mIoU and 0.6\% OA on NYU-Depth v2. 

Promisingly, consistent improvements are observed by simply transferring the pretrained backbones to SOTA supervised mutimodal fusion models. Following the published codebase and without tuning any hyperparameter, we push CMX-B2 from 49.7\% to 50.6\% in mIoU on the SUN-RGBD dataset, and CMX-B5 from 56.9\% to 57.3\% in mIoU on the NYU-Depth v2 dataset.

\begin{table}[t]
\caption{RGB-depth fine-tuning results on SUN-RGBD and NYU-Depth v2. 
}
\label{tab:rgbd-mm}
\centering
\scalebox{0.75}{
\begin{tabular}{lcccc}
\toprule
{SUN-RGBD} & Modal     & mIoU      & OA      \\ \hline
FCN \cite{long2015fully}          & RGB  & 27.4      & 68.2      \\
FCN (CLIP \cite{radford2021learning}) & RGB & 30.5 & 74.2 \\
FCN (DeCUR)          & RGB  & 34.5      & 75.5      \\ \hdashline
SA-Gate \cite{chen2020bi}      & RGBD & 49.4           & 82.5      \\
SGNet \cite{chen2021spatial}        & RGBD & 48.6           & 82.0      \\
ShapeConv \cite{cao2021shapeconv}    & RGBD & 48.6           & 82.2      \\ \hdashline
CMX-B2 \cite{zhang2022cmx}       & RGBD & 49.7         & 82.8      \\
CMX-B2 (DeCUR)       & RGBD & \textbf{50.6} & \textbf{83.2} \\ \bottomrule
\end{tabular}
}
\scalebox{0.75}{
\begin{tabular}{lcccc}
\toprule
{NYUDv2} & modal     & mIoU      & OA    \\ \hline
FCN \cite{long2015fully}          & RGB  & 29.2      & 60.0      \\
FCN (CLIP \cite{radford2021learning}) & RGB & 30.4  & 63.3  \\
FCN (DeCUR)          & RGB  & 31.2      & 63.9      \\ \hdashline
SA-Gate \cite{chen2020bi}      & RGBD & 52.4           & 77.9      \\
ShapeConv \cite{cao2021shapeconv}    & RGBD & 51.3           & 76.4      \\ 
OMNIVORE \cite{Girdhar_2022_CVPR}    & {RGBD} & {54.0}           & {-} \\
\hdashline
CMX-B5 \cite{zhang2022cmx}       & RGBD & 56.9         & 80.1      \\
CMX-B5 (DeCUR)       & RGBD & \textbf{57.3} & \textbf{80.3} \\ \bottomrule
\end{tabular}
}
\end{table}

\section{Ablation studies} \label{sec:abexp}


\subsubsection{Deformable attention} 

We conduct frozen-encoder ablation on DeCUR for the deformable attention with residual connection (RDA) as shown in \cref{tab:ablation-rda}. While DA without residual connection decreases label-limited performance, the RDA module consistently improves in almost all scenarios.

\begin{table}[htp]
\caption{Ablation results (mAP) on the deformable attention module.}
\label{tab:ablation-rda}
\centering
\scalebox{0.8}{
\begin{tabular}{c|cc|cc|cc|cc}
\toprule
         \multirow{2}{*}{Dataset} & \multicolumn{2}{c|}{BigEarthNet-MM} & \multicolumn{2}{c|}{BigEarthNet-SAR} & \multicolumn{2}{c|}{GeoNRW-MM} & \multicolumn{2}{c}{GeoNRW-RGB} \\
         & 1\%             & 100\%            & 1\%              & 100\%            & 1\%          & 100\%         & 1\%          & 100\%          \\ \hline
w/o. DA & 79.4            & 85.4             & 73.7             & 78.3             & 34.9          & 43.9          & 31.4          & 38.4           \\
with DA   & $-$0.1            & -                & $-$0.2             & -                & $-$0.6          & -             & $-$1.2          & -              \\
with RDA & $+$0.4            & $+$0.8             & $+$0.7             & $+$1.2             & $-$0.2          & $+$0.8          & $+$0.8          & $+$2.4           \\ \bottomrule
\end{tabular}
}
\end{table}

\subsubsection{Loss terms}  
The components of the DeCUR loss are ablated in \cref{tab:ablation-loss}, where fine-tuning is conducted without the RDA module. As shown, when neither intra-modal nor modality-unique information is learned, the performance of the single cross-modal Barlow Twins (w/o intra\&decoup.) significantly degrades in downstream tasks. Additionally, solely incorporating modality decoupling (w/o intra) yields unstable effects in the two scenarios. This can be explained by the fact that without intra-modal training the unique dimensions can be randomly decorrelated and are not ensured meaningful. When intra-modal training is added (w/o decoup.), the model's performance improves consistently in both scenarios, albeit inferior to DeCUR. This underscores the complementary benefits of intra-modal representations to commonly learned cross-modal representations. Together, these results affirm the effectiveness of the total DeCUR loss.

\begin{table}[htp]
\caption{Ablation results on loss components, where \textit{intra} and \textit{decoup.} correspond to intra-modal training and modality decoupling, respectively. }
\label{tab:ablation-loss}
\centering
\scalebox{0.8}{
\begin{tabular}{lcc}
\toprule
\multicolumn{1}{c}{} &  SAR-optical (mAP)        & RGB-DEM (mIoU)             \\ \hline
DeCUR (ours)     & \textbf{81.7} & \textbf{36.9} \\
w/o intra\&decoup.     & 80.3                & 33.6                \\
w/o intra              & 80.1                & 34.3                \\
w/o decoup.          & 81.1                & 35.2                \\ \bottomrule
\end{tabular}
}
\end{table}

\subsubsection{Decoupling percentage} 
\begin{figure}[h]
\centering
\begin{subfigure}{0.5\linewidth}
\includegraphics[width=0.95\linewidth]{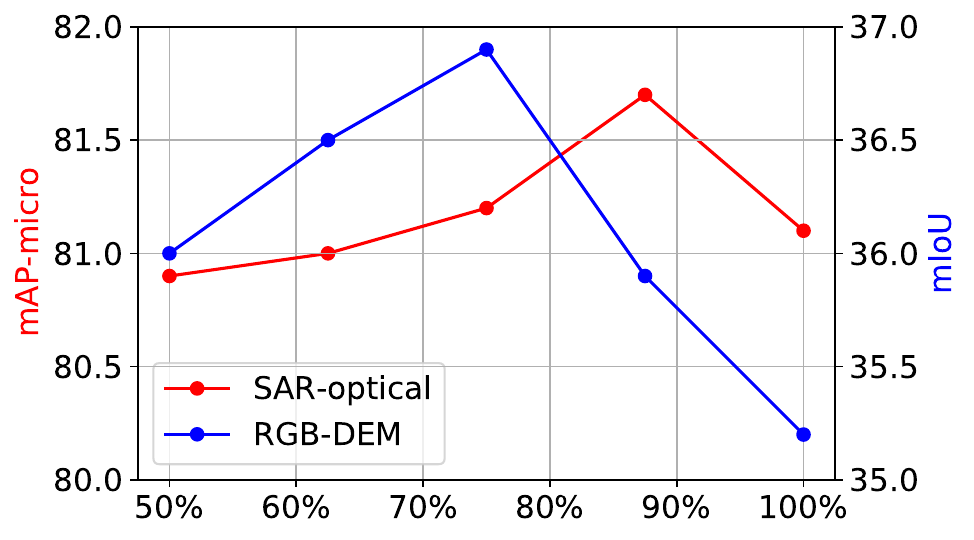}
\caption{Ablation results on the percentage of common dimensions.
}
\label{fig:ablation-cu}
\end{subfigure}
\begin{subfigure}{0.45\linewidth}
\includegraphics[width=0.95\linewidth]{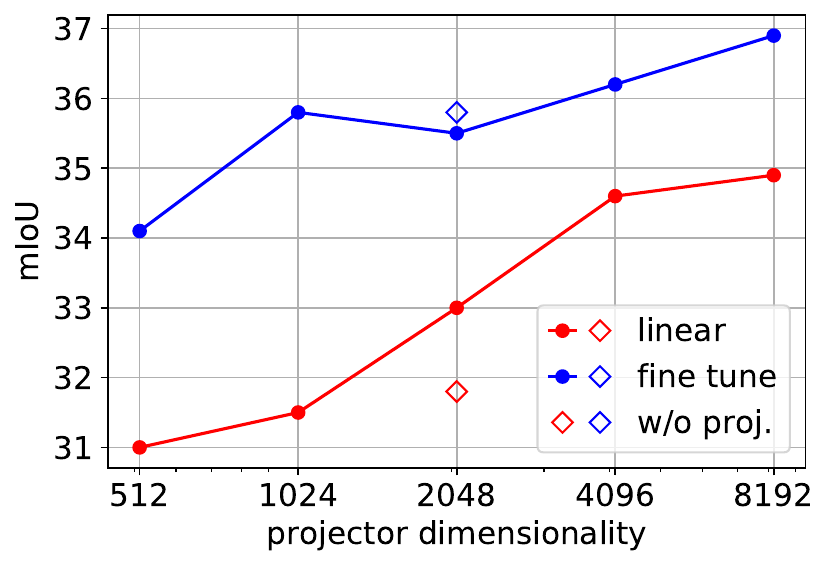}
\caption{Effect of the projector dimensionality on the GeoNRW dataset.
}
\label{fig:ablation-prj}
\end{subfigure}
\caption{Ablation results on the percentage of common dimensions and the projector.}
\end{figure}

We conduct a simple grid search to find the best ratio between common and unique dimensions for different modality combinations that may have different representation overlaps. As shown in \cref{fig:ablation-cu}, the best percentage of common dimensions is 87.5\% for SAR-optical and 75\% for RGB-DEM. This could be in line with the fact that there is more valid modality-unique information in orthophoto and elevation models than in optical and SAR (when the optical image is almost cloud-free). In both scenarios, the downstream performance increases and decreases smoothly along with the change of the percentage of common dimensions. Interestingly, there is no significant performance drop when decoupling up to 50\% unique dimensions, which indicates the sparsity of the common embedding space.


\subsubsection{Effect of the projector}  
Inherited from Barlow Twins \cite{zbontar2021barlow}, DeCUR also benefits from the increasing dimensionality of the projector. As can be seen in \cref{fig:ablation-prj}, DeCUR keeps improving with all output dimensionality tested.
Interestingly, DeCUR works well on the segmentation task even without the projector. Removing the projector gives reasonable downstream performances, while adding it back can further enhance the representations. 

\subsubsection{Robustness of decoupling percentage}

\begin{figure}[htp]
\centering
\includegraphics[width=0.43\linewidth]{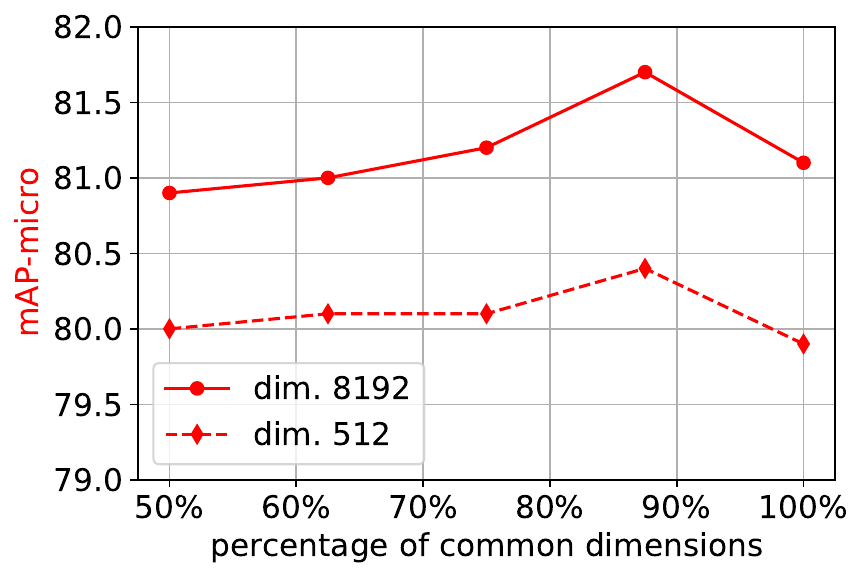}
\includegraphics[width=0.42\linewidth]{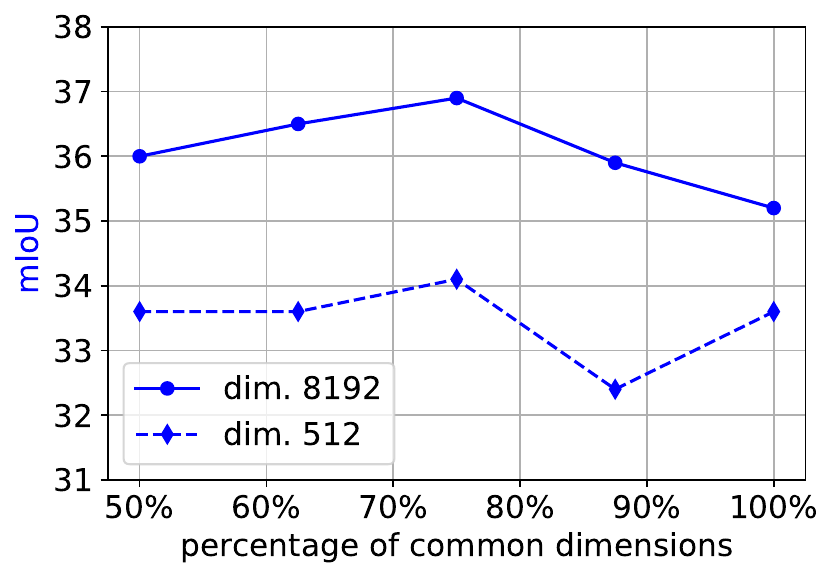}
\caption{Ablation on the decoupling percentage with different embedding dimensionalities (left: SAR-optical; right: RGB-DEM).
}
\label{fig:ablation-cu2}
\end{figure}

The grid search in \cref{fig:ablation-cu} was built upon an embedding dimension of 8192. To see how the best percentage changes along with the embedding dimensionality, we repeat the search with an embedding dimension of 512 on SAR-optical and RGB-DEM datasets.
As is shown in \cref{fig:ablation-cu2}, the best percentage of common dimensions is interestingly the same for both the small and the big embedding spaces. 

\section{Discussion}
In this section, we demonstrate an explainability analysis to interpret the multimodal representations learned by DeCUR. We illustrate SAR-optical analysis on the SSL4EO-S12 dataset as an example scenario here.

\subsubsection{Cross-modal representation alignment} 
\begin{figure}[htp]
\centering
\begin{subfigure}{0.52\linewidth}
\includegraphics[width=0.95\linewidth]{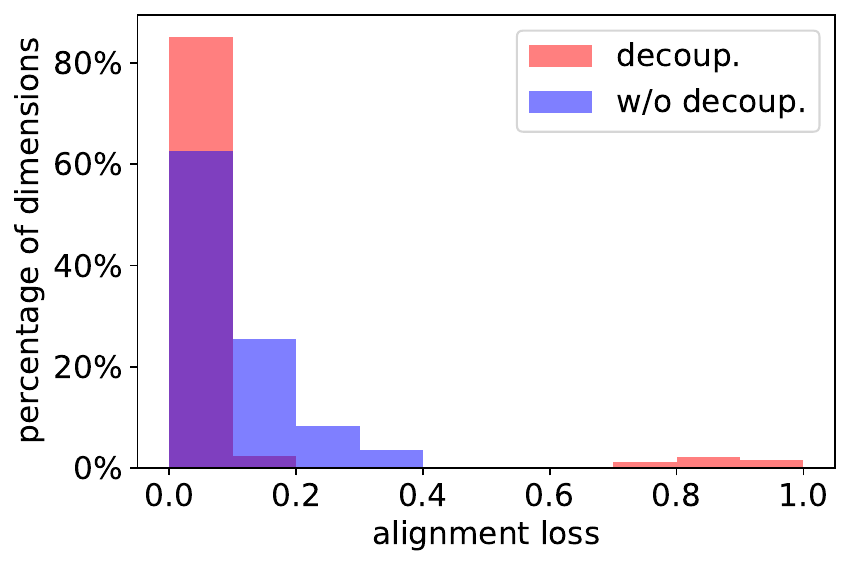}
\caption{Representation alignment analysis.}
\label{fig:align_BE}
\end{subfigure}
\begin{subfigure}{0.42\linewidth}
\includegraphics[width=0.9\linewidth]{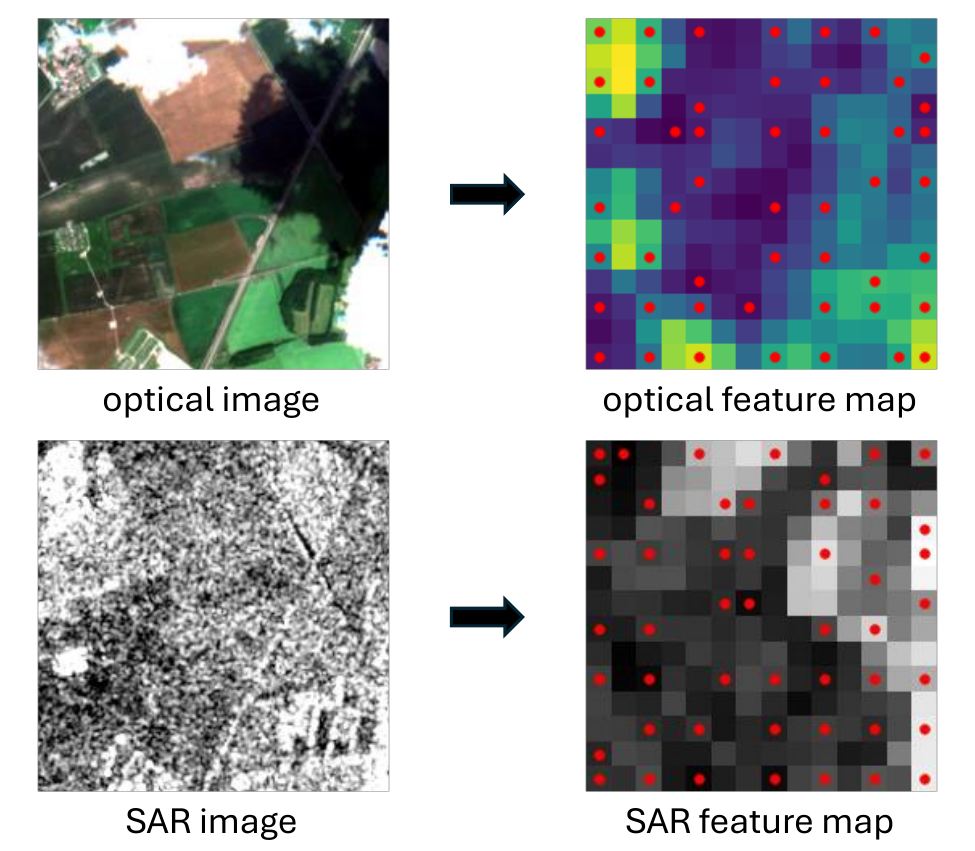}
\caption{Feature maps with deformable points.}
\label{fig:vis_rda}
\end{subfigure}
\caption{Cross-modal representation alignment (left) and DA visualization (right).}
\end{figure}

To examine that each modality contains unique information that is difficult to integrate into a common space, we calculate the cross-modal alignment of every embedding dimension. This is done by counting the on-diagonal losses of the cross-correlation matrix $\mathcal{C}$:
\begin{equation}
\mathcal{L}_{i} = \left(1-\mathcal{C}_{i i}\right)^2  \quad,
\end{equation}
\noindent where $i$ is the $i_{th}$ embedding dimension. The closer $\mathcal{L}_{i}$ to 0, the better the alignment of the two modalities in this dimension. We count the loss for all dimensions and plot the histogram of one random batch for both DeCUR and cross-modal Barlow Twins. 
The former explicitly decouples unique dimensions, while the latter assumes that all dimensions are common. As shown in \cref{fig:align_BE}, the alignment loss remains high for a certain number of dimensions with cross-modal Barlow Twins. On the contrary, by allowing the decorrelation of several dimensions (the loss of which moves to 1), the misalignment of common dimensions decreases. 
These results are aligned with the visualization by t-SNE mentioned above in \cref{fig:decur-p1}, where modality-unique dimensions are well separated, while common dimensions are perfectly overlapped in DeCUR.

\subsubsection{Deformable feature visualization}
We visualize the first principle component \cite{dunteman1989principal} of the feature maps with deformable points learned by the deformable attention module in \cref{fig:vis_rda}. 
For illustration purposes, we visualize in color for optical and in grey for SAR. It shows that the two modalities make different focuses in this scene: the model learns to ignore clouds in the optical image, while on the contrary paying more attention to such regions in the SAR image as radar signal can go through the clouds. The deformable points in general follow the feature attention, but spread more sparsely compared to natural vision. This is due to two facts: 1) satellite imagery is not object-centric and information is spread out, and 2) the residual connection restricts the deforming capacity.

\subsubsection{Spatial and spectral saliency statistics}  We count spatial and spectral saliency statistics over the whole dataset. For preparation, 
we average the common and unique dimensions as two single values. Next, one-time backpropagation is performed w.r.t the corresponding output target (0 for common and 1 for unique) to get GradCAM \cite{selvaraju2017grad} and Integrated Gradients \cite{sundararajan2017axiomatic}. One "common" and one "unique" saliency map are generated for each modality with both visualization methods. 

\begin{figure}[htp]
\centering
\begin{subfigure}{0.43\linewidth}
\includegraphics[width=\linewidth]{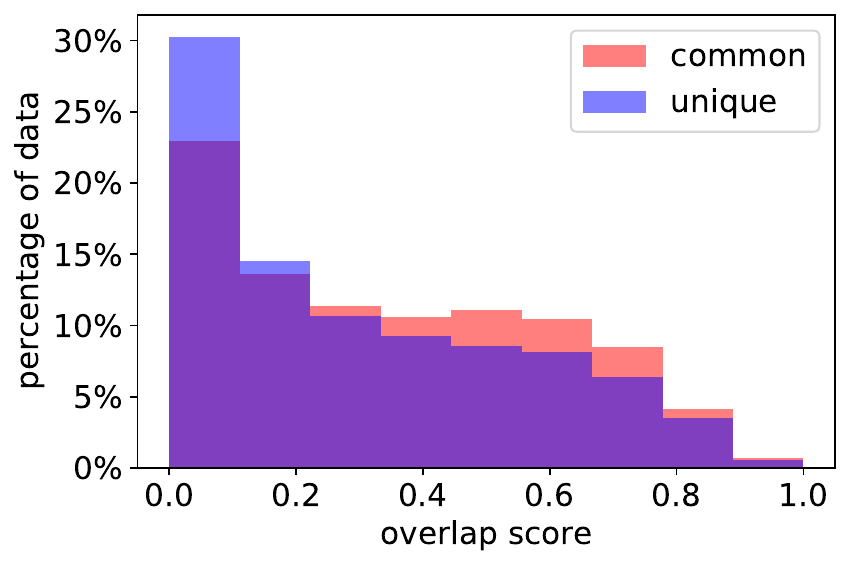}
\caption{SAR-optical spatial saliency statistics.}
\label{fig:stat_BE_spatial}
\end{subfigure}
\begin{subfigure}{0.55\linewidth}
\includegraphics[width=0.87\linewidth]{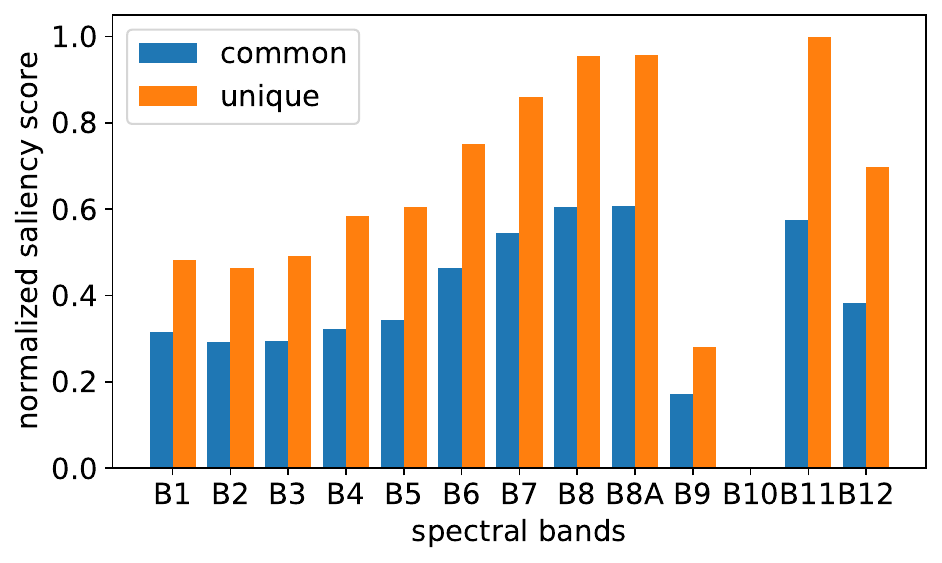}
\caption{Spectral saliency statistics for 13 optical bands.}
\label{fig:stat_BE_band_S2}
\end{subfigure}
\caption{Spatial saliency statistics (left) and spectral saliency statistics (right).}
\label{fig:stat_BE}
\end{figure}

Then, we use GradCAM saliency maps to count spatial statistics, calculating one overlap score for the common area and one for the unique area between the modalities, respectively. Going through the whole dataset, we can get a histogram in \cref{fig:stat_BE_spatial}. The histogram shows a trend of unique scores being more towards 0 than common scores, indicating that the interesting areas of modality-unique representations tend to be more orthogonal than common representations which tend to overlap. 

On the other hand, We average the importance scores of each band in the Integrated Gradients saliency maps to get spectral saliency for both common and unique representations. We normalize the scores, count statistics over the whole dataset, and plot the histograms in \cref{fig:stat_BE_band_S2}. The figure confirms the bigger influence of the spectral information on optical-unique representations. Meanwhile, the band-wise importance distribution is promisingly consistent with the domain knowledge: 1) near-infrared bands (B5-B8A, including vegetation red edge) are very important; 2) water vapor (B9) and cirrus (B10) are strongly related to the atmosphere and thus less important for land surface monitoring; etc.

\section{Conclusion}
We propose DeCUR, a multimodal self-supervised learning method that learns both cross-modal common and modality-unique representations. 
Extensive experiments on three common multimodal scenarios prove the effectiveness of DeCUR, suggesting the great potential of modality-decoupling. 

One limitation of DeCUR is it simplifies the multimodal situation by allocating the same common-unique ratio across the dataset. Future work could consider more complex scenarios where one modality may contain more unique information than the other in different scenes.
Another limitation is the grid search for the best percentage of common dimensions, which can be costly on a huge dataset. While a general percentage of around 80\% can achieve reasonable performance in our tested scenarios, a more efficient discovering strategy is to be explored. Other directions for future work include the exploration of adaptive decoupling strategies, and integrating modality decoupling in unified foundation models with more than two modalities.



\section*{Acknowledgement}
The main work of Y.\ Wang, C.\ Liu, and C.\ Albrecht was funded by the Helmholtz Association through the Framework of Helmholtz AI, grant ID: \texttt{ZT-I-PF-5-01} - \textit{Local Unit Munich Unit @Aeronautics, Space and Transport (MASTr)}. The compute was supported by the Helmholtz Association's Initiative and Networking Fund on the HAICORE@FZJ partition. The work of N. Ait Ali Braham was supported by the European Commission through the project "EvoLand" under the Horizon 2020 Research and Innovation program (Grant Agreement No. 101082130). The work of X. Zhu was supported by the German Federal Ministry of Education and Research (BMBF) in the framework of the international future AI lab "AI4EO -- Artificial Intelligence for Earth Observation: Reasoning, Uncertainties, Ethics and Beyond" (grant number: 01DD20001) and by the Munich Center for Machine Learning. The work of Z. Xiong was supported by the German Federal Ministry for the Environment, Nature Conservation, Nuclear Safety and Consumer Protection (BMUV) based on a resolution of the German Bundestag (grant number: 67KI32002B; Acronym: \textit{EKAPEx}). Y. Wang's work on rebuttal and camera-ready paper preparation was supported by the European Commission through the project “ThinkingEarth—Copernicus Foundation Models for a Thinking Earth” under the Horizon 2020 Research and Innovation program (Grant Agreement No. 101130544).


%
%
\bibliographystyle{splncs04}
\bibliography{main}

\begin{thebibliography}{10}
\providecommand{\url}[1]{\texttt{#1}}
\providecommand{\urlprefix}{URL }
\providecommand{\doi}[1]{https://doi.org/#1}

\bibitem{akbari2021vatt}
Akbari, H., Yuan, L., Qian, R., Chuang, W.H., Chang, S.F., Cui, Y., Gong, B.: Vatt: Transformers for multimodal self-supervised learning from raw video, audio and text. Advances in Neural Information Processing Systems  \textbf{34},  24206--24221 (2021)

\bibitem{s5xq-b822-20}
Baier, G., Deschemps, A., Schmitt, M., Yokoya, N.: Geonrw (2020). \doi{10.21227/s5xq-b822}, \url{https://dx.doi.org/10.21227/s5xq-b822}

\bibitem{bardes2021vicreg}
Bardes, A., Ponce, J., LeCun, Y.: Vicreg: Variance-invariance-covariance regularization for self-supervised learning. arXiv preprint arXiv:2105.04906  (2021)

\bibitem{cao2021shapeconv}
Cao, J., Leng, H., Lischinski, D., Cohen-Or, D., Tu, C., Li, Y.: Shapeconv: Shape-aware convolutional layer for indoor rgb-d semantic segmentation. In: Proceedings of the IEEE/CVF international conference on computer vision. pp. 7088--7097 (2021)

\bibitem{caron2020unsupervised}
Caron, M., Misra, I., Mairal, J., Goyal, P., Bojanowski, P., Joulin, A.: Unsupervised learning of visual features by contrasting cluster assignments. Advances in neural information processing systems  \textbf{33},  9912--9924 (2020)

\bibitem{caron2021emerging}
Caron, M., Touvron, H., Misra, I., J{\'e}gou, H., Mairal, J., Bojanowski, P., Joulin, A.: Emerging properties in self-supervised vision transformers. In: Proceedings of the IEEE/CVF international conference on computer vision. pp. 9650--9660 (2021)

\bibitem{chen2021spatial}
Chen, L.Z., Lin, Z., Wang, Z., Yang, Y.L., Cheng, M.M.: Spatial information guided convolution for real-time rgbd semantic segmentation. IEEE Transactions on Image Processing  \textbf{30},  2313--2324 (2021)

\bibitem{chen2020simple}
Chen, T., Kornblith, S., Norouzi, M., Hinton, G.: A simple framework for contrastive learning of visual representations. In: International conference on machine learning. pp. 1597--1607. PMLR (2020)

\bibitem{chen2020bi}
Chen, X., Lin, K.Y., Wang, J., Wu, W., Qian, C., Li, H., Zeng, G.: Bi-directional cross-modality feature propagation with separation-and-aggregation gate for rgb-d semantic segmentation. In: European Conference on Computer Vision. pp. 561--577. Springer (2020)

\bibitem{chen2021exploring}
Chen, X., He, K.: Exploring simple siamese representation learning. In: Proceedings of the IEEE/CVF conference on computer vision and pattern recognition. pp. 15750--15758 (2021)

\bibitem{cong2022satmae}
Cong, Y., Khanna, S., Meng, C., Liu, P., Rozi, E., He, Y., Burke, M., Lobell, D., Ermon, S.: Satmae: Pre-training transformers for temporal and multi-spectral satellite imagery. Advances in Neural Information Processing Systems  \textbf{35},  197--211 (2022)

\bibitem{dunteman1989principal}
Dunteman, G.H.: Principal components analysis, vol.~69. Sage (1989)

\bibitem{ericsson2022self}
Ericsson, L., Gouk, H., Loy, C.C., Hospedales, T.M.: Self-supervised representation learning: Introduction, advances, and challenges. IEEE Signal Processing Magazine  \textbf{39}(3),  42--62 (2022)

\bibitem{feng2023cross}
Feng, Z., Song, L., Yang, S., Zhang, X., Jiao, L.: Cross-modal contrastive learning for remote sensing image classification. IEEE Transactions on Geoscience and Remote Sensing  (2023)

\bibitem{fuller2024croma}
Fuller, A., Millard, K., Green, J.: Croma: Remote sensing representations with contrastive radar-optical masked autoencoders. Advances in Neural Information Processing Systems  \textbf{36} (2024)

\bibitem{gidaris2018unsupervised}
Gidaris, S., Singh, P., Komodakis, N.: Unsupervised representation learning by predicting image rotations. arXiv preprint arXiv:1803.07728  (2018)

\bibitem{Girdhar_2023_CVPR}
Girdhar, R., El-Nouby, A., Liu, Z., Singh, M., Alwala, K.V., Joulin, A., Misra, I.: Imagebind: One embedding space to bind them all. In: Proceedings of the IEEE/CVF Conference on Computer Vision and Pattern Recognition (CVPR). pp. 15180--15190 (June 2023)

\bibitem{Girdhar_2022_CVPR}
Girdhar, R., Singh, M., Ravi, N., van~der Maaten, L., Joulin, A., Misra, I.: Omnivore: A single model for many visual modalities. In: Proceedings of the IEEE/CVF Conference on Computer Vision and Pattern Recognition (CVPR). pp. 16102--16112 (June 2022)

\bibitem{grill2020bootstrap}
Grill, J.B., Strub, F., Altch{\'e}, F., Tallec, C., Richemond, P., Buchatskaya, E., Doersch, C., Avila~Pires, B., Guo, Z., Gheshlaghi~Azar, M., et~al.: Bootstrap your own latent-a new approach to self-supervised learning. Advances in neural information processing systems  \textbf{33},  21271--21284 (2020)

\bibitem{guo2023skysense}
Guo, X., Lao, J., Dang, B., Zhang, Y., Yu, L., Ru, L., Zhong, L., Huang, Z., Wu, K., Hu, D., et~al.: Skysense: A multi-modal remote sensing foundation model towards universal interpretation for earth observation imagery. arXiv preprint arXiv:2312.10115  (2023)

\bibitem{gupta2014learning}
Gupta, S., Girshick, R., Arbel{\'a}ez, P., Malik, J.: Learning rich features from rgb-d images for object detection and segmentation. In: Computer Vision--ECCV 2014: 13th European Conference, Zurich, Switzerland, September 6-12, 2014, Proceedings, Part VII 13. pp. 345--360. Springer (2014)

\bibitem{he2022masked}
He, K., Chen, X., Xie, S., Li, Y., Doll{\'a}r, P., Girshick, R.: Masked autoencoders are scalable vision learners. In: Proceedings of the IEEE/CVF Conference on Computer Vision and Pattern Recognition. pp. 16000--16009 (2022)

\bibitem{he2020momentum}
He, K., Fan, H., Wu, Y., Xie, S., Girshick, R.: Momentum contrast for unsupervised visual representation learning. In: Proceedings of the IEEE/CVF conference on computer vision and pattern recognition. pp. 9729--9738 (2020)

\bibitem{he2016deep}
He, K., Zhang, X., Ren, S., Sun, J.: Deep residual learning for image recognition. In: Proceedings of the IEEE conference on computer vision and pattern recognition. pp. 770--778 (2016)

\bibitem{hong2023spectralgpt}
Hong, D., Zhang, B., Li, X., Li, Y., Li, C., Yao, J., Yokoya, N., Li, H., Jia, X., Plaza, A., et~al.: Spectralgpt: Spectral foundation model. arXiv preprint arXiv:2311.07113  (2023)

\bibitem{jha2023gaf}
Jha, A., Bose, S., Banerjee, B.: Gaf-net: improving the performance of remote sensing image fusion using novel global self and cross attention learning. In: Proceedings of the IEEE/CVF Winter Conference on Applications of Computer Vision. pp. 6354--6363 (2023)

\bibitem{krishnan2022self}
Krishnan, R., Rajpurkar, P., Topol, E.J.: Self-supervised learning in medicine and healthcare. Nature Biomedical Engineering  \textbf{6}(12),  1346--1352 (2022)

\bibitem{liang2024factorized}
Liang, P.P., Deng, Z., Ma, M.Q., Zou, J.Y., Morency, L.P., Salakhutdinov, R.: Factorized contrastive learning: Going beyond multi-view redundancy. Advances in Neural Information Processing Systems  \textbf{36} (2024)

\bibitem{long2015fully}
Long, J., Shelhamer, E., Darrell, T.: Fully convolutional networks for semantic segmentation. In: Proceedings of the IEEE conference on computer vision and pattern recognition. pp. 3431--3440 (2015)

\bibitem{loshchilov2016sgdr}
Loshchilov, I., Hutter, F.: Sgdr: Stochastic gradient descent with warm restarts. arXiv preprint arXiv:1608.03983  (2016)

\bibitem{van2008visualizing}
Van~der Maaten, L., Hinton, G.: Visualizing data using t-sne. Journal of machine learning research  \textbf{9}(11) (2008)

\bibitem{manas2021seasonal}
Manas, O., Lacoste, A., Gir{\'o}-i Nieto, X., Vazquez, D., Rodriguez, P.: Seasonal contrast: Unsupervised pre-training from uncurated remote sensing data. In: Proceedings of the IEEE/CVF International Conference on Computer Vision. pp. 9414--9423 (2021)

\bibitem{mendieta2023towards}
Mendieta, M., Han, B., Shi, X., Zhu, Y., Chen, C.: Towards geospatial foundation models via continual pretraining. In: Proceedings of the IEEE/CVF International Conference on Computer Vision. pp. 16806--16816 (2023)

\bibitem{mu2022slip}
Mu, N., Kirillov, A., Wagner, D., Xie, S.: Slip: Self-supervision meets language-image pre-training. In: European Conference on Computer Vision. pp. 529--544. Springer Nature Switzerland Cham (2022)

\bibitem{Silberman:ECCV12}
Nathan~Silberman, Derek~Hoiem, P.K., Fergus, R.: Indoor segmentation and support inference from rgbd images. In: ECCV (2012)

\bibitem{oord2018representation}
Oord, A.v.d., Li, Y., Vinyals, O.: Representation learning with contrastive predictive coding. arXiv preprint arXiv:1807.03748  (2018)

\bibitem{peng2022balanced}
Peng, X., Wei, Y., Deng, A., Wang, D., Hu, D.: Balanced multimodal learning via on-the-fly gradient modulation. In: Proceedings of the IEEE/CVF Conference on Computer Vision and Pattern Recognition. pp. 8238--8247 (2022)

\bibitem{radford2021learning}
Radford, A., Kim, J.W., Hallacy, C., Ramesh, A., Goh, G., Agarwal, S., Sastry, G., Askell, A., Mishkin, P., Clark, J., et~al.: Learning transferable visual models from natural language supervision. In: International conference on machine learning. pp. 8748--8763. PMLR (2021)

\bibitem{scheibenreif2022self}
Scheibenreif, L., Hanna, J., Mommert, M., Borth, D.: Self-supervised vision transformers for land-cover segmentation and classification. In: Proceedings of the IEEE/CVF Conference on Computer Vision and Pattern Recognition. pp. 1422--1431 (2022)

\bibitem{selvaraju2017grad}
Selvaraju, R.R., Cogswell, M., Das, A., Vedantam, R., Parikh, D., Batra, D.: Grad-cam: Visual explanations from deep networks via gradient-based localization. In: Proceedings of the IEEE international conference on computer vision. pp. 618--626 (2017)

\bibitem{song2015sun}
Song, S., Lichtenberg, S.P., Xiao, J.: Sun rgb-d: A rgb-d scene understanding benchmark suite. In: Proceedings of the IEEE conference on computer vision and pattern recognition. pp. 567--576 (2015)

\bibitem{sumbul2021bigearthnet}
Sumbul, G., De~Wall, A., Kreuziger, T., Marcelino, F., Costa, H., Benevides, P., Caetano, M., Demir, B., Markl, V.: Bigearthnet-mm: A large-scale, multimodal, multilabel benchmark archive for remote sensing image classification and retrieval [software and data sets]. IEEE Geoscience and Remote Sensing Magazine  \textbf{9}(3),  174--180 (2021)

\bibitem{sundararajan2017axiomatic}
Sundararajan, M., Taly, A., Yan, Q.: Axiomatic attribution for deep networks. In: International conference on machine learning. pp. 3319--3328. PMLR (2017)

\bibitem{vincent2010stacked}
Vincent, P., Larochelle, H., Lajoie, I., Bengio, Y., Manzagol, P.A., Bottou, L.: Stacked denoising autoencoders: Learning useful representations in a deep network with a local denoising criterion. Journal of machine learning research  \textbf{11}(12) (2010)

\bibitem{wang2021multimodal}
Wang, L., Luc, P., Recasens, A., Alayrac, J.B., Oord, A.v.d.: Multimodal self-supervised learning of general audio representations. arXiv preprint arXiv:2104.12807  (2021)

\bibitem{wang2022self}
Wang, Y., Albrecht, C.M., Braham, N.A.A., Mou, L., Zhu, X.X.: Self-supervised learning in remote sensing: A review. arXiv preprint arXiv:2206.13188  (2022)

\bibitem{wang2022self1}
Wang, Y., Albrecht, C.M., Zhu, X.X.: Self-supervised vision transformers for joint sar-optical representation learning. In: IGARSS 2022-2022 IEEE International Geoscience and Remote Sensing Symposium. pp. 139--142. IEEE (2022)

\bibitem{wang2022ssl4eo}
Wang, Y., Braham, N.A.A., Xiong, Z., Liu, C., Albrecht, C.M., Zhu, X.X.: {SSL4EO-S12}: A large-scale multi-modal, multi-temporal dataset for self-supervised learning in earth observation. arXiv preprint arXiv:2211.07044  (2022)

\bibitem{wang2023ssl4eo}
Wang, Y., Braham, N.A.A., Xiong, Z., Liu, C., Albrecht, C.M., Zhu, X.X.: Ssl4eo-s12: A large-scale multimodal, multitemporal dataset for self-supervised learning in earth observation [software and data sets]. IEEE Geoscience and Remote Sensing Magazine  \textbf{11}(3),  98--106 (2023)

\bibitem{wang2023feature}
Wang, Y., Hern{\'a}ndez, H.H., Albrecht, C.M., Zhu, X.X.: Feature guided masked autoencoder for self-supervised learning in remote sensing. arXiv preprint arXiv:2310.18653  (2023)

\bibitem{wei2022mvp}
Wei, L., Xie, L., Zhou, W., Li, H., Tian, Q.: Mvp: Multimodality-guided visual pre-training. In: European Conference on Computer Vision. pp. 337--353. Springer Nature Switzerland Cham (2022)

\bibitem{woo2018cbam}
Woo, S., Park, J., Lee, J.Y., Kweon, I.S.: Cbam: Convolutional block attention module. In: Proceedings of the European conference on computer vision (ECCV). pp. 3--19 (2018)

\bibitem{xia2022vision}
Xia, Z., Pan, X., Song, S., Li, L.E., Huang, G.: Vision transformer with deformable attention. In: Proceedings of the IEEE/CVF conference on computer vision and pattern recognition. pp. 4794--4803 (2022)

\bibitem{xia2023dat++}
Xia, Z., Pan, X., Song, S., Li, L.E., Huang, G.: Dat++: Spatially dynamic vision transformer with deformable attention. arXiv preprint arXiv:2309.01430  (2023)

\bibitem{xie2021segformer}
Xie, E., Wang, W., Yu, Z., Anandkumar, A., Alvarez, J.M., Luo, P.: Segformer: Simple and efficient design for semantic segmentation with transformers. Advances in Neural Information Processing Systems  \textbf{34},  12077--12090 (2021)

\bibitem{xiong2024neural}
Xiong, Z., Wang, Y., Zhang, F., Stewart, A.J., Hanna, J., Borth, D., Papoutsis, I., Saux, B.L., Camps-Valls, G., Zhu, X.X.: Neural plasticity-inspired foundation model for observing the earth crossing modalities. arXiv preprint arXiv:2403.15356  (2024)

\bibitem{xiong2024one}
Xiong, Z., Wang, Y., Zhang, F., Zhu, X.X.: One for all: Toward unified foundation models for earth vision. arXiv preprint arXiv:2401.07527  (2024)

\bibitem{xiong2020msn}
Xiong, Z., Yuan, Y., Wang, Q.: Msn: Modality separation networks for rgb-d scene recognition. Neurocomputing  \textbf{373},  81--89 (2020)

\bibitem{xiong2021ask}
Xiong, Z., Yuan, Y., Wang, Q.: Ask: Adaptively selecting key local features for rgb-d scene recognition. IEEE Transactions on Image Processing  \textbf{30},  2722--2733 (2021)

\bibitem{yang2022vision}
Yang, J., Duan, J., Tran, S., Xu, Y., Chanda, S., Chen, L., Zeng, B., Chilimbi, T., Huang, J.: Vision-language pre-training with triple contrastive learning. In: Proceedings of the IEEE/CVF Conference on Computer Vision and Pattern Recognition. pp. 15671--15680 (2022)

\bibitem{you2017large}
You, Y., Gitman, I., Ginsburg, B.: Large batch training of convolutional networks. arXiv preprint arXiv:1708.03888  (2017)

\bibitem{zbontar2021barlow}
Zbontar, J., Jing, L., Misra, I., LeCun, Y., Deny, S.: Barlow twins: Self-supervised learning via redundancy reduction. In: International Conference on Machine Learning. pp. 12310--12320. PMLR (2021)

\bibitem{zhang2022cmx}
Zhang, J., Liu, H., Yang, K., Hu, X., Liu, R., Stiefelhagen, R.: Cmx: Cross-modal fusion for rgb-x semantic segmentation with transformers. arXiv preprint arXiv:2203.04838  (2022)

\bibitem{zhou2023feature}
Zhou, J., Yu, Q., Luo, C., Zhang, J.: Feature decomposition for reducing negative transfer: A novel multi-task learning method for recommender system. arXiv preprint arXiv:2302.05031  (2023)

\end{thebibliography}


\appendix


\section{Algorithm}

\begin{algorithm}[]
\SetAlgoLined
\footnotesize
    \PyComment{f1, f2: encoder networks} \\
    \PyComment{BN, N, K: batch normalization, batch size and embedding dimension} \\
    \PyComment{on\_diag, off\_diag: on- and off-diagonal elements of a matrix} \\
    \vspace{0.5em}
    \PyComment{loss function for common and intra-modal} \\
    \PyCode{def loss\_c(C, lambda):} \\
    \Indp
        \PyCode{l\_on = (on\_diag(C)-1).pow(2).sum()} \\
        \PyCode{l\_off = off\_diag(C).pow(2).sum()} \\
        \PyCode{return l\_on + lambda x l\_off} \\
    \Indm
    \vspace{0.5em}
    \PyComment{loss function for unique} \\
    \PyCode{def loss\_u(C, lambda):} \\
    \Indp
        \PyCode{l\_on = on\_diag(C).pow(2).sum()} \\
        \PyCode{l\_off = off\_diag(C).pow(2).sum()} \\
        \PyCode{return l\_on + lambda x l\_off} \\
    \Indm
    \vspace{0.5em}
    \PyComment{training} \\
    \PyCode{for x1,x2 in loader:} \PyComment{load a batch pairs} \\
    \Indp   
        \PyCode{(x1\_1, x1\_2), (x2\_1, x2\_2) = augment1(x1), augment2(x2)}  \\ 
        \PyComment{compute embeddings and normalize} \\
        \PyCode{z1\_1, z1\_2 = BN(f1(x1\_1)), BN(f1(x1\_2))}  \\ 
        \PyCode{z2\_1, z2\_2 = BN(f2(x2\_1)), BN(f2(x2\_2))}  \\ 
        \PyComment{cross-correlation matrices} \\
        \PyCode{C1 = z1\_1.T @ z1\_2 / N} \PyComment{KxK} \\ 
        \PyCode{C2 = z2\_1.T @ z2\_2 / N} \PyComment{KxK} \\
        \PyCode{Cm = z1\_1.T @ z2\_1 / N} \PyComment{KxK} \\
        \PyCode{Cc = Cm[:K\_c,:K\_c]} \PyComment{KcxKc} \\
        \PyCode{Cu = Cm[K\_c:,K\_c:]} \PyComment{KuxKu} \\
        \PyComment{calculate losses} \\
        \PyCode{L1 = loss\_c(C1,lmb1)} \PyComment{intra-modal M1} \\
        \PyCode{L2 = loss\_c(C2,lmb2)} \PyComment{intra-modal M2} \\
        \PyCode{Lc = loss\_c(Cc,lmbc)} \PyComment{cross-m. common} \\
        \PyCode{Lu = loss\_u(Cu,lmbu)} \PyComment{cross-m. unique} \\
        \PyCode{loss = L1 + L2 + Lc + Lu} \PyComment{total loss} \\        
        \PyComment{optimization} \\
        \PyCode{loss.backward()} \\
        \PyCode{optimizer.step()} \\
    \Indm 

\caption{PyTorch-style pseudocode for DeCUR.}
\label{algo:decur}
\end{algorithm}

\section{Implementation Details}
\label{app:implementation}
\subsection{Pretraining}
\subsubsection{SAR-optical pretraining}
\hspace{0.5em}
SSL4EO-S12 \cite{wang2022ssl4eo} dataset is used for SAR-optical pretraining: Sentinel-1 GRD (2 bands VV and VH) and Sentinel-2 L1C (13 multispectral bands). The pixel resolution is united to 10 meters. We compress and normalize the optical data to 8-bit by band-wise mean and standard deviation; for SAR data, we cut out 2\% outliers for each image and normalize it by band-wise mean and standard deviation. 

Standard ResNet50 is used as the encoder backbone, of which the first layer is modified to fit the input channel number. The projector is a 3-layer MLP, of which the first two layers include Linear, BactchNorm and ReLU, and the last one includes only a linear layer. We adopt two residual deformable attention (RDA) modules after the last two blocks of the ResNet encoder. Following \cite{xia2023dat++}, for the first RDA module, we use a feature map 14$\times$14, 8 heads with 128 channels each, 4 groups, stride 2, and kernel size 5; for the second RDA module, we use a feature map 7$\times$7, 16 heads with 128 channels each, 8 groups, stride 1, and kernel size 3.

We use the LARS \cite{you2017large} optimizer with weight decay 1e-6 and momentum 0.9. We use a learning rate of 0.2 for the weights and 0.0048 for the biases and batch normalization parameters. We reduce the learning rate using a cosine decay schedule \cite{loshchilov2016sgdr} (no warm-up periods). The biases and batch normalization parameters are excluded from LARS adaptation and weight decay.

\subsubsection{RGB-DEM pretraining}
\hspace{0.5em}
The training split of the GeoNRW \cite{s5xq-b822-20} dataset is used for RGB-DEM pretraining: aerial orthophoto (3 bands RGB) and lidar-derived digital elevation model (1 band height). The pixel resolution is 1 meter.
We use standard ResNet50 without modifying the input layer (i.e., we duplicate the DEM image to 3 channels). Other model architecture and optimization protocols are the same as SAR-optical pretraining.

\subsubsection{RGB-depth pretraining}
\hspace{0.5em}
SUN-RGBD \cite{song2015sun} dataset is used for RGB-depth pretraining: indoor RGB and depth images. Following \cite{zhang2022cmx}, we preprocess the depth images to HHA format \cite{gupta2014learning}. We use standard ResNet50 and MiT-B2/B5 from SegFormer as the backbones. For segformer backbones, we use AdamW optimizer and a learning rate of 1e-4.

\subsubsection{Data augmentations} 
We follow common augmentations in the self-supervision literature \cite{grill2020bootstrap} for optical and RGB images (resized crop, color jitter, grayscale, Gaussian blur, horizontal and vertical flip, color drop, solarize), and remove infeasible ones for special modalities. Specifically, for SAR, we use random resized crop, grayscale, Gaussian blur, and horizontal and vertical flip; for DEM images, we use random resized crop and horizontal and vertical flip; for HHA images, we use random resized crop and horizontal flip.

\subsection{Transfer learning}
\subsubsection{SAR-optical transfer learning}
\hspace{0.5em}
We evaluate SAR-optical pretraining on the BigEarthNet-MM \cite{sumbul2021bigearthnet} dataset for the multi-label scene classification task. We compress and normalize the optical images to 8-bit by band-wise mean and standard deviation; for SAR images, we cut out 2\% outliers for each image and normalize it by band-wise mean and standard deviation. As the optical data of BigEarthNet-MM is Sentinel-2 L2A product (12 bands), we insert one empty band to match the pretrained weights (13 bands). We use common data augmentations including RandomResizedCrop (scale 0.8 to 1) and RandomHorizontalFlip.

Standard ResNet50 is used as the encoder backbone for each modality, of which the first layer is modified to fit the input channel number, and the last layer is modified as an identity layer. The encoded features are concatenated, followed by a fully connected layer outputting the class logits. The encoders are initialized from the pretrained weights. For linear classification, the encoder weights are frozen and only the last classification layer is trainable; for fine-tuning, all weights are trained.

We optimize MultiLabelSoftMarginLoss with batch size 256 for 100 epochs. We use the SGD optimizer with a weight decay of 0 and momentum of 0.9. The learning rate is 0.5 for linear classification, and 0.05 for fine-tuning. We reduce the learning rate by factor 10 at 60 and 80 epochs.

\subsubsection{RGB-DEM transfer learning}
\hspace{0.5em}
We evaluate RGB-DEM pretraining on the GeoNRW dataset for the semantic segmentation task. We use common data augmentations including RandomResizedCrop (scale 0.2 to 1) and RandomHorizontalFlip.

Fully convolutional networks (FCN) \cite{long2015fully} with standard ResNet50 backbone for each modality are used as the segmentation model. The last three feature maps from both modalities are concatenated and upsampled to the input size. They are further followed by 1x1 convolution outputting three segmentation maps, which are added together to form the final output map. The encoders are initialized from the pretrained weights. For linear classification, the encoder weights are frozen; for fine-tuning, all weights are trainable.

We optimize CrossEntropyLoss with batch size 256 for 30 epochs. We use the AdamW optimizer with a weight decay of 0.01. The learning rate is 0.0001 for both linear classification and fine-tuning. 

\subsubsection{RGB-depth transfer learning}
\hspace{0.5em}
We evaluate RGB-depth pretraining on SUN-RGBD and NYU-Depth v2 datasets for the semantic segmentation task. We use common data augmentations including RandomResizedCrop and RandomHorizontalFlip.

FCN with ResNet50 backbones is used as the segmentation model for single-modal RGB semantic segmentation. We optimize CrossEntropyLoss with batch size 8 for 40k iterations. We use the SGD optimizer with weight decay 1e-5. The learning rate is 0.01 with polynomial decay for fine-tuning.

CMX \cite{zhang2022cmx} with segformer \cite{xie2021segformer} backbones are used as the segmentation model for RGBD semantic segmentation. We follow the same settings of CMX for SUN-RGBD and NYU-depth v2 datasets.

\section{Additional results}

\subsubsection{Additional results with ViT backbones}
\hspace{0.5em}
To complement experiments with ResNet backbones in the main paper, we provide in \cref{tab:vits} additional frozen-encoder results with ViT backbones on SAR-optical and RGB-DEM scenarios. The consistent improvement of DeCUR over SimCLR verifies again its architecture-agnostic property.

\begin{table}[h]
    \centering
    \caption{Additional frozen-encoder results with ViT backbones.}
    \label{tab:vits}
\begin{tabular}{llcc}
\toprule
& Backbone             & BigEarthNet-SAR-1\% & GeoNRW-RGB-1\% \\ \hline
Rand. Init. & ViT-S/16 & 57.7      & 13.2           \\
SimCLR & ViT-S/16 &   73.9         & 35.1           \\
DeCUR (ours) & ViT-S/16 & \textbf{75.8}  & \textbf{35.6}  \\ \bottomrule
\end{tabular}

\end{table}

\vspace{-1.5em}
\subsubsection{Additional results with attention mechanisms}
\hspace{0.5em}
To further support the benefits of adopting deformable attention, we provide in \cref{tab:attention} additional frozen-encoder results with different attention designs. Specifically, our design is better than the popular convolutional block attention module (CBAM) \cite{woo2018cbam}. In addition, we also explore the possibility of cross-modal attention: taking queries from one modality, and keys and values from another modality, which results in performance decreasing. While this cross-attention tends to be beneficial in supervised learning \cite{jha2023gaf,feng2023cross}, The reason could be a potential information leak that hurts the self-supervised pretraining task.

\vspace{-1em}
\begin{table}[h]
    \centering
    \caption{Additional frozen-encoder results with different attention designs. \textit{DA} means deformable attention without residual connection; \textit{Cross DA} represents cross-modal deformation attention; \textit{RDA} represents deformable attention with residual connection.}
    \label{tab:attention}
\begin{tabular}{lcc}
\toprule
             & BigEarthNet-SAR-1\% & GeoNRW-RGB-1\% \\ \hline
DA \cite{xia2022vision}  & 73.5    & 30.2 \\
Cross DA     & 72.5                  & 29.6                       \\
CBAM \cite{woo2018cbam} & 73.1                       & 31.0                       \\
RDA (ours)   & \textbf{74.4}          & \textbf{32.2}              \\ \bottomrule
\end{tabular}

\end{table}

\vspace{-1.5em}
\section{Explainability analysis}

\subsection{Algorithm}
For a better understanding of our explainability implementation, we provide united pseudocode of 1) cross-modal representation alignment, 2) t-SNE representation visualization, 3) spatial and spectral saliency statistics based on GradCam and Integrated Gradients, in Algorithm \ref{algo:explain}.

\begin{algorithm}[]
\footnotesize
\SetAlgoLined
    \PyComment{f1,f2: encoder networks} \\
    \PyComment{BN: batch normalization} \\
    \PyComment{N,K: batch size and embedding dimension} \\
    \PyComment{on\_diag: on-diagonal elements of a matrix} \\
    \PyComment{IG: Integrated Gradients} \\
    \vspace{1em}
    \PyComment{1.Cross-modal representation alignment} \\
    \PyCode{def alignment\_histogram(z1, z2):} \\
    \Indp
        \PyCode{C = z1.T @ z2 / N} \PyComment{KxK} \\
        \PyCode{losses = (on\_diag(C)-1).pow(2)} \PyComment{Kx1} \\
        \PyCode{return histogram(losses, range=(0,1))} \\
    \Indm
    \vspace{1em}
    \PyComment{2.Representation visualization} \\
    \PyCode{def tsne\_vis(z1, z2):} \\
    \Indp
        \PyCode{feature = torch.cat((z1,z2),-1)} \PyComment{Nx2K} \\
        \PyCode{feature = feature.permute(1,0)} \PyComment{2KxN} \\
        \PyCode{return tsne(feature, n\_components=2)} \\
    \Indm
    \vspace{1em}
    \PyComment{3.Spatial saliency visualization} \\
    \PyCode{def gradcam\_vis(x1):} \\
    \Indp
        \PyCode{z1 = BN(f1(x1))} \PyComment{NxK} \\
        \PyCode{z1\_c = z1[:,:Kc].mean(dim=-1)} \PyComment{Nx1} \\
        \PyCode{z1\_u = z1[:,Kc:].mean(dim=-1)} \PyComment{Nx1} \\
        \PyCode{out1 = torch.cat((z1\_c,z1\_u),-1)} \PyComment{Nx2} \\
        \PyCode{gc1 = LayerGradCam(f1, f1.last\_conv2)} \\
        \PyCode{attr1\_c = gc1.attribute(x1,target=0)} \PyComment{Nx7x7} \\
        \PyCode{attr1\_u = gc1.attribute(x1,target=1)} \PyComment{Nx7x7} \\
        \PyCode{return upsamp(attr1\_c), upsamp(attr1\_u)} \\
        \PyComment{Nx224x224, Nx224x224} \\        
    \Indm    
    \vspace{1em}
    \PyComment{4.Spatial saliency statistics} \\
    \PyCode{def gradcam\_stat(x1,x2):} \\
    \Indp
        \PyCode{att1\_c, att1\_u = gradcam\_vis(x1)} \\
        \PyCode{att2\_c, att2\_u = gradcam\_vis(x2)} \\
        \PyCode{mul\_c = norm(att1\_c) x norm(att2\_c)} \PyComment{Nx1} \\
        \PyCode{mul\_u = norm(att1\_u) x norm(att2\_u)} \PyComment{Nx1} \\
        return mul\_c, mul\_u \\
    \Indm
    \vspace{1em}
    \PyComment{5.Spectral saliency statistics} \\
    \PyCode{def IG\_stat(x1):} \\
    \Indp
        \PyComment{define "IG\_vis" similar to gradcam} \\
        \PyCode{att1\_c, att1\_u = IG\_vis(x1)} \PyComment{NxCx224x224} \\
        \PyCode{imp\_c = att1\_c.mean(dim=(0,2,3))} \PyComment{NxC} \\
        \PyCode{imp\_u = att1\_u.mean(dim=(0,2,3))} \PyComment{NxC} \\
        return imp\_c, imp\_u \\
    \Indm

\caption{Pseudocode for DeCUR explainability.}
\label{algo:explain}
\end{algorithm}

\subsection{Additional examples}

Below we show some additional explainability examples. Note that the decoupling and matching results depend on the samples. Specifically, some images have a strong overlap between modalities (potentially more common dimensions) while others tend to be more orthogonal (potentially more unique dimensions, decoupling helps more).

\vspace{-1em}
\subsubsection{Cross-modal alignment histograms} 
\begin{figure}[]
\centering
   \includegraphics[width=0.45\linewidth]{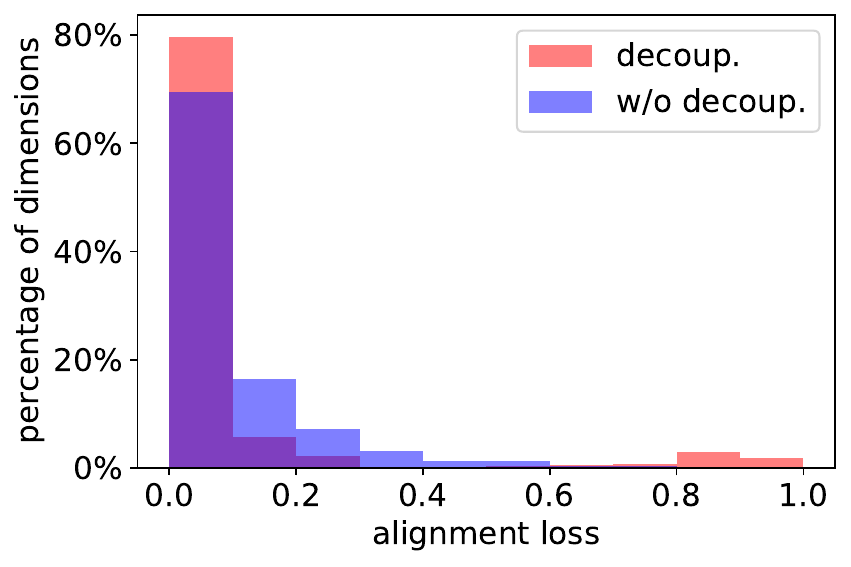}
   \includegraphics[width=0.45\linewidth]{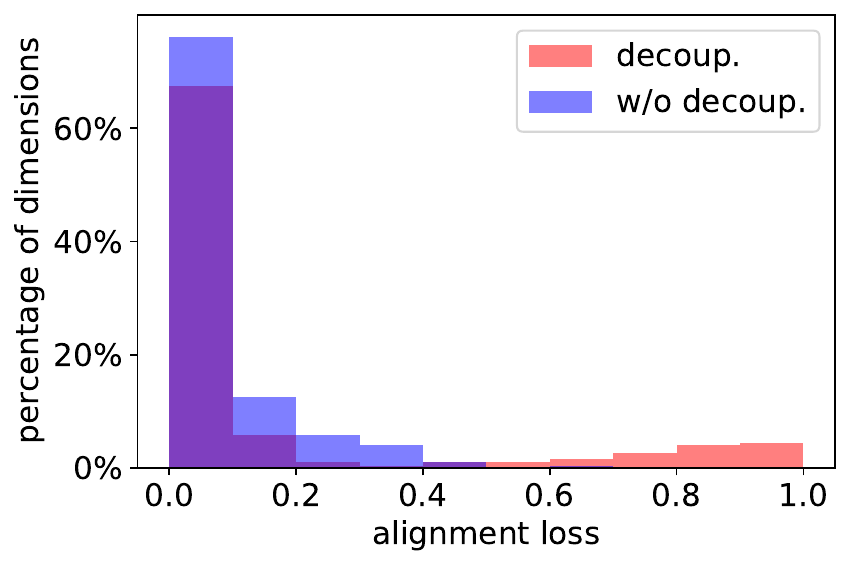}
   \includegraphics[width=0.45\linewidth]{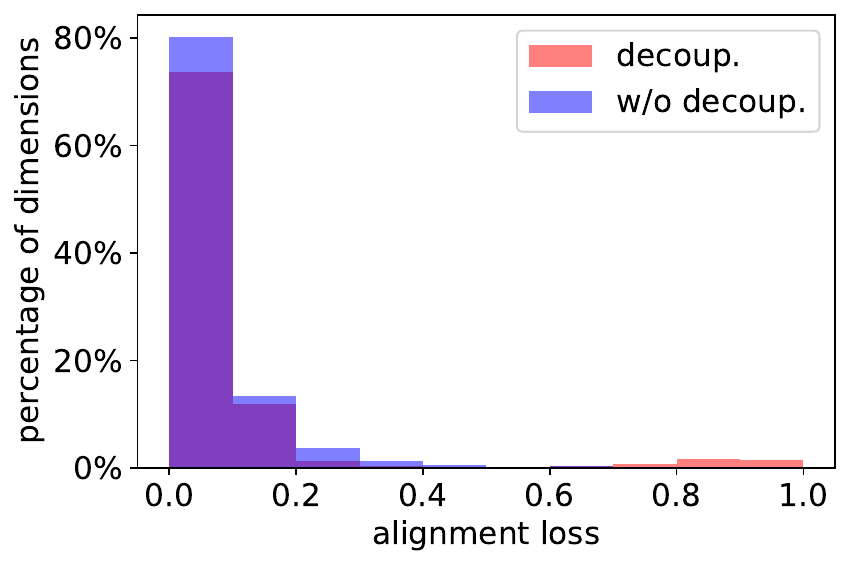}
   \includegraphics[width=0.45\linewidth]{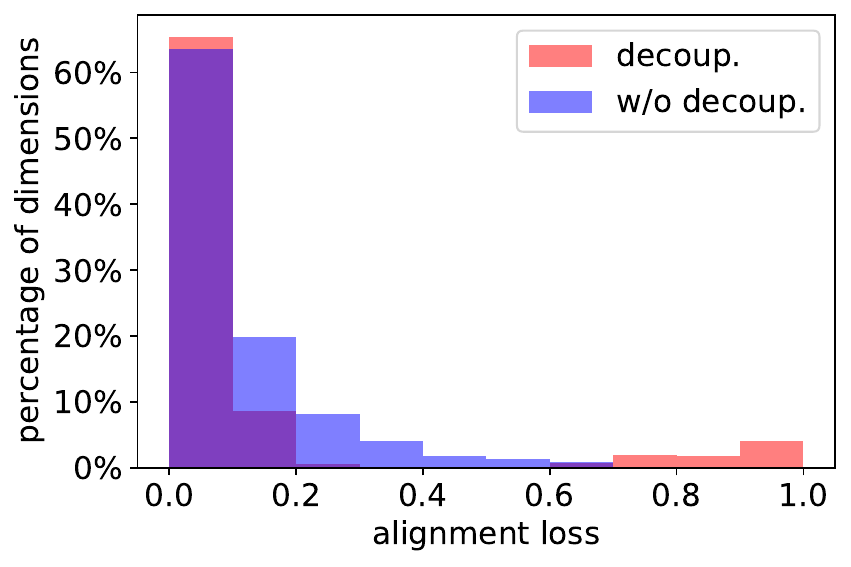}
   \includegraphics[width=0.45\linewidth]{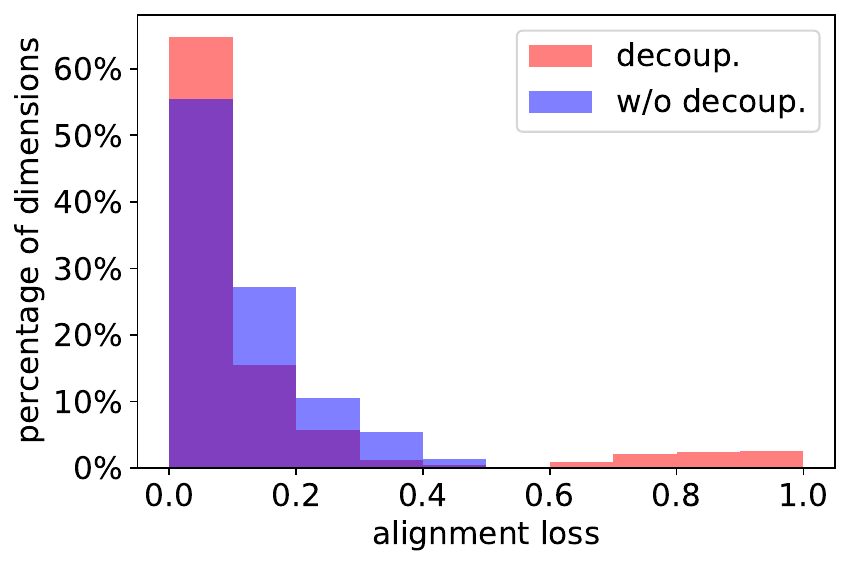}
   \includegraphics[width=0.45\linewidth]{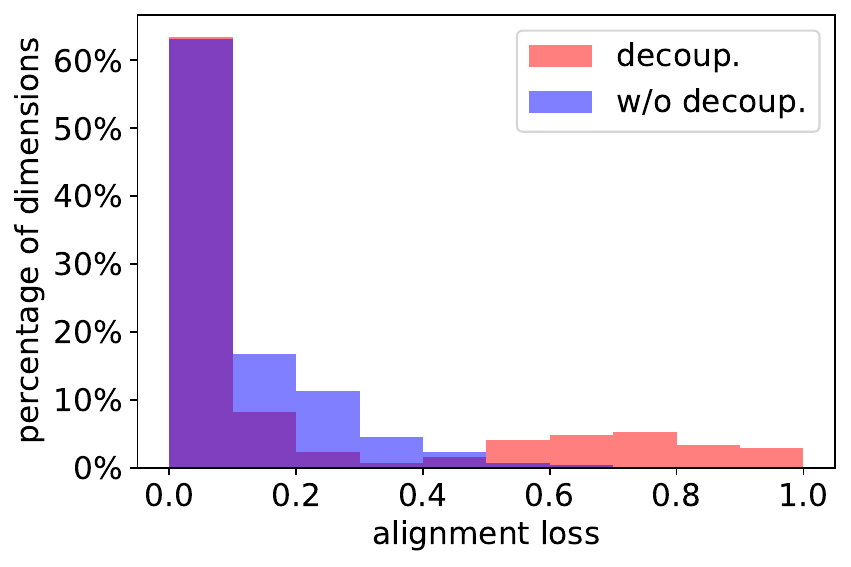}
   \includegraphics[width=0.45\linewidth]{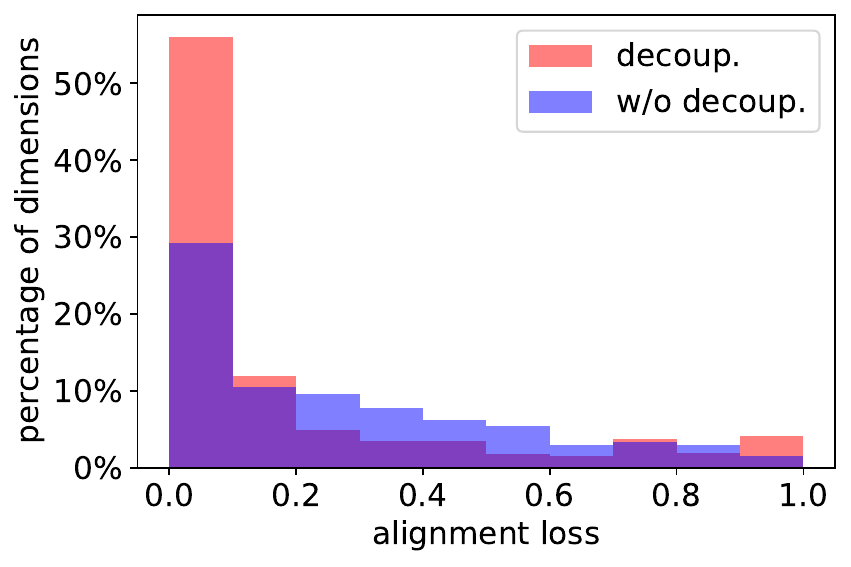}
   \includegraphics[width=0.45\linewidth]{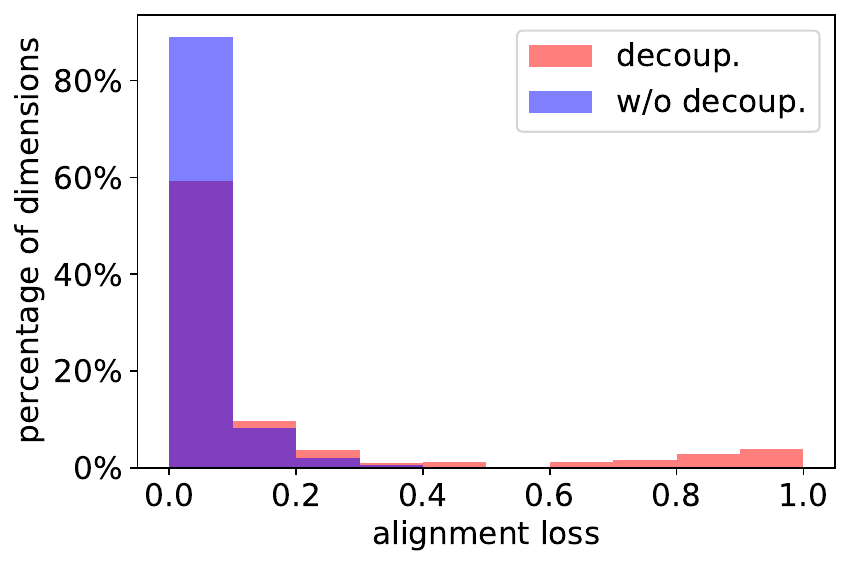}
   \caption{Cross-modal representation alignment histograms of 4 batches of samples. Left: SAR-optical; right: RGB-DEM.}
\label{fig:align_more}
\end{figure}

See Figure \ref{fig:align_more}.

\vspace{-1em}
\subsubsection{t-SNE representation visualization} 
\begin{figure}[]
\centering
   \includegraphics[width=0.48\linewidth]{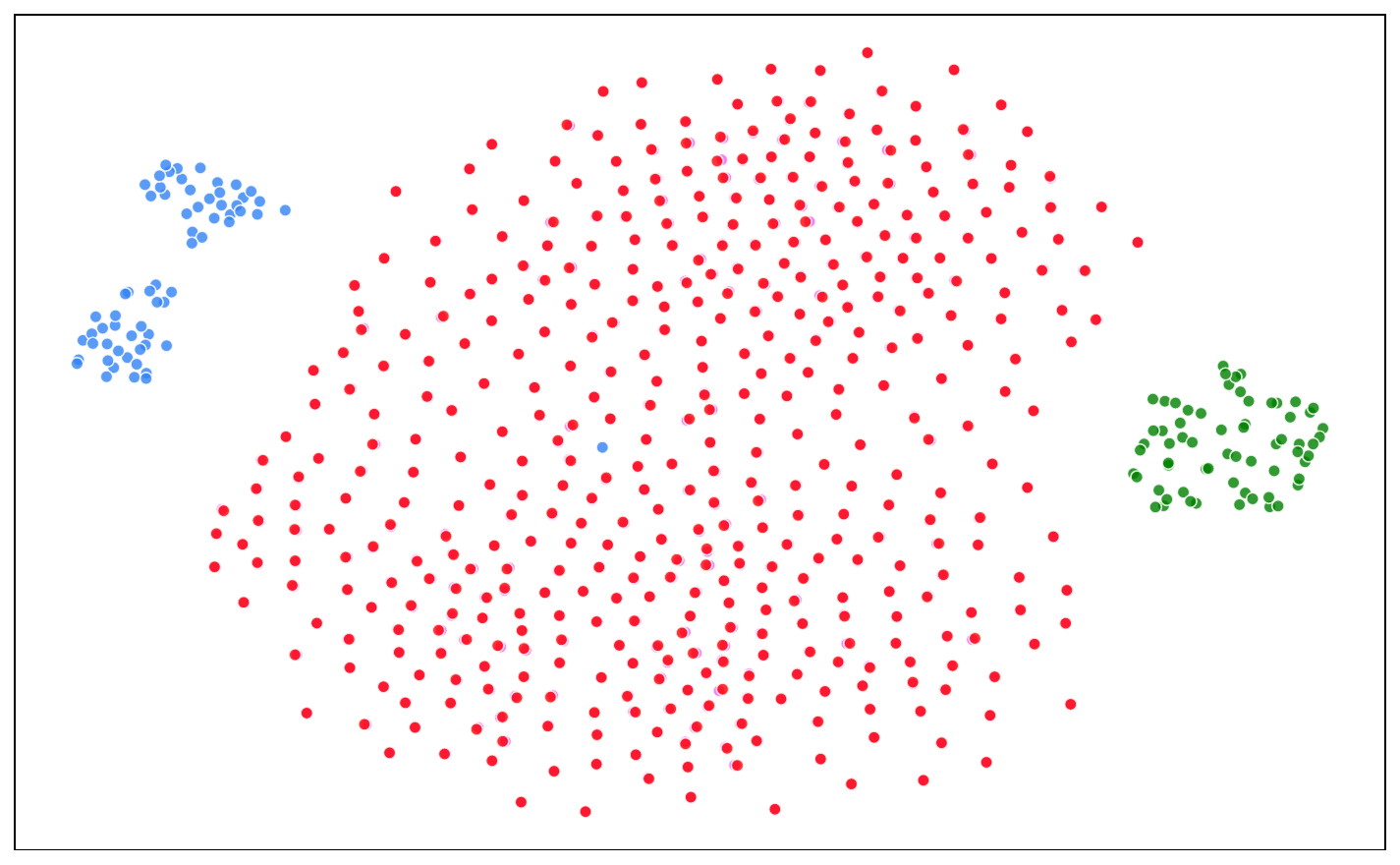}
   \includegraphics[width=0.48\linewidth]{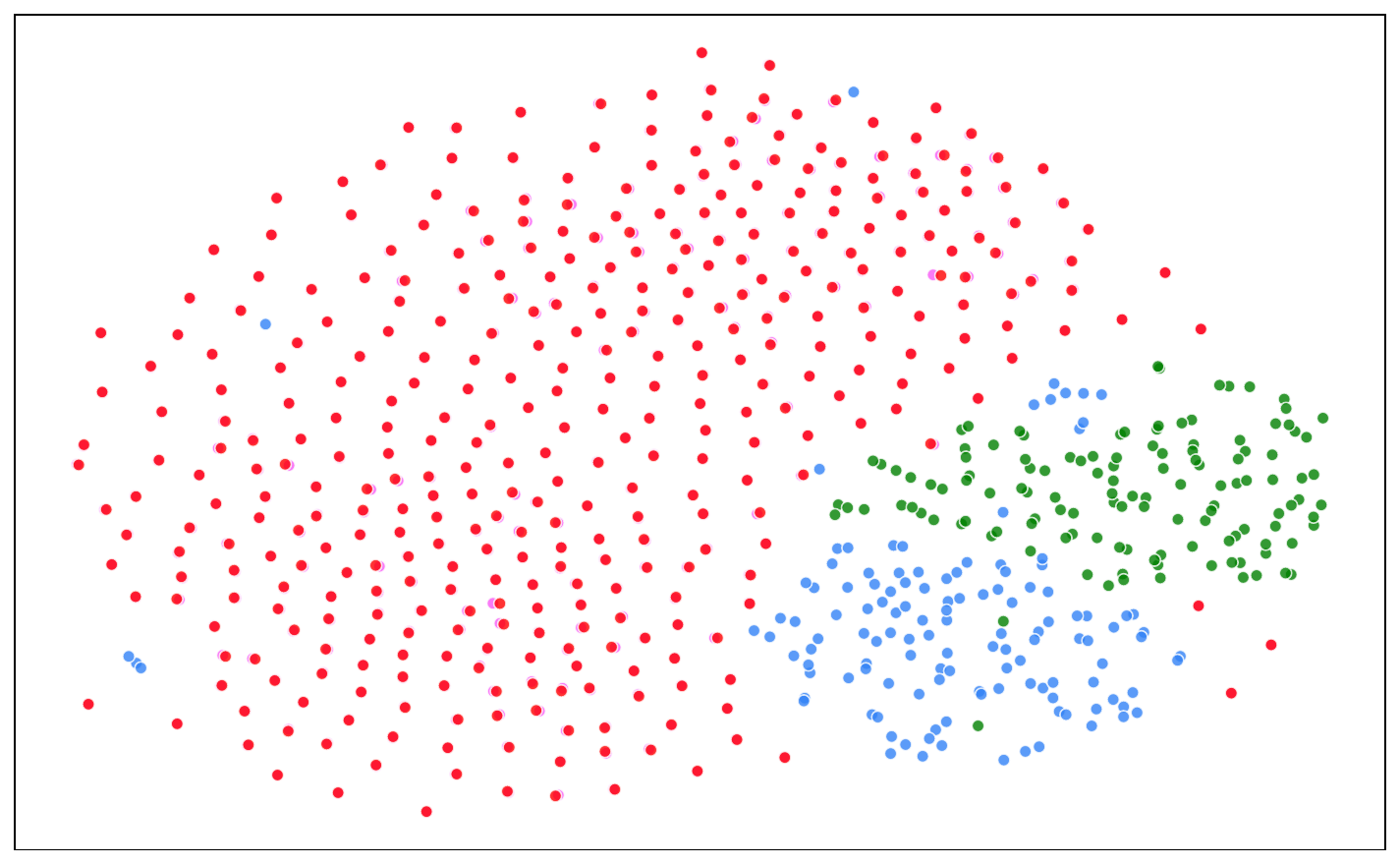}
   \includegraphics[width=0.48\linewidth]{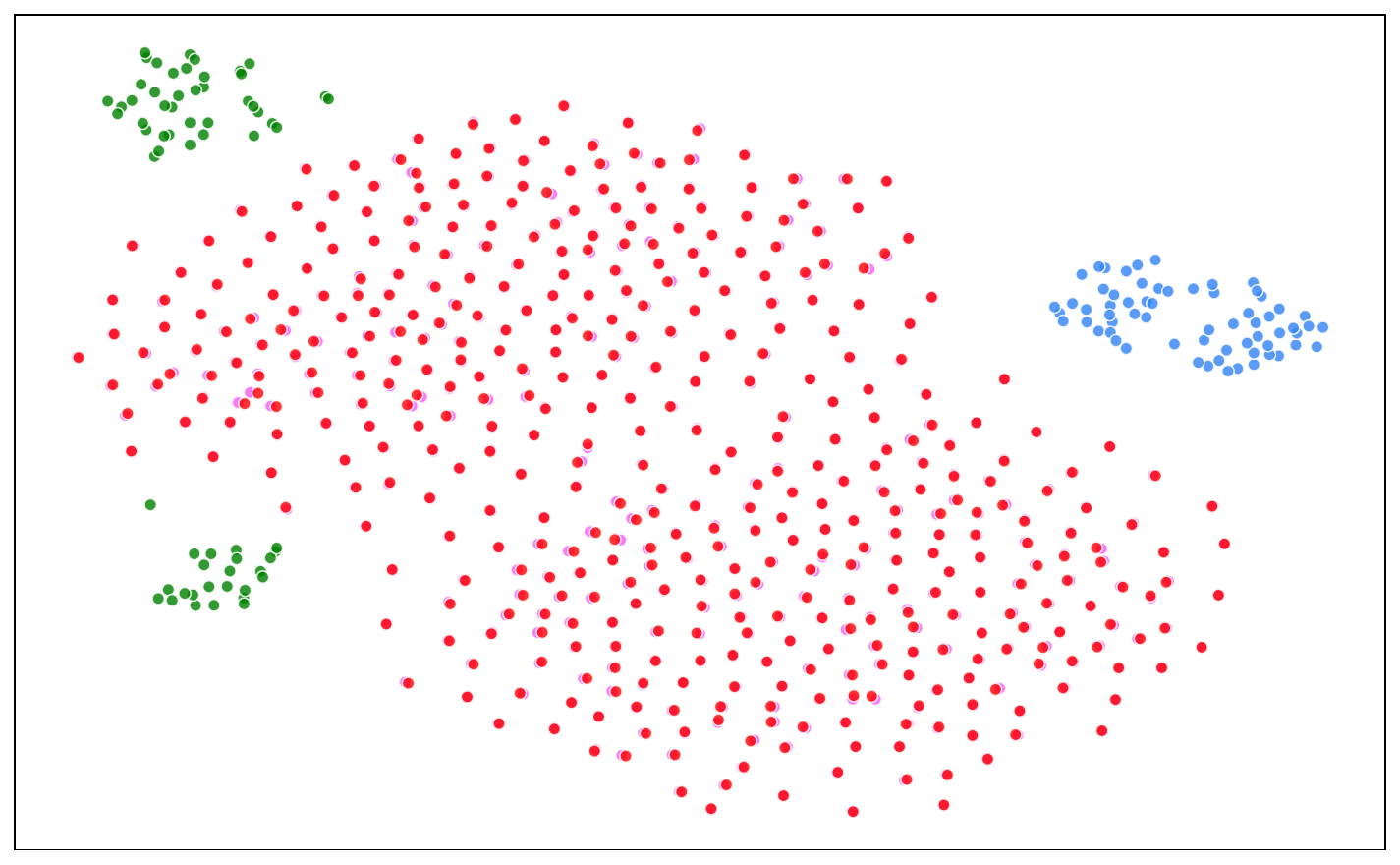}
   \includegraphics[width=0.48\linewidth]{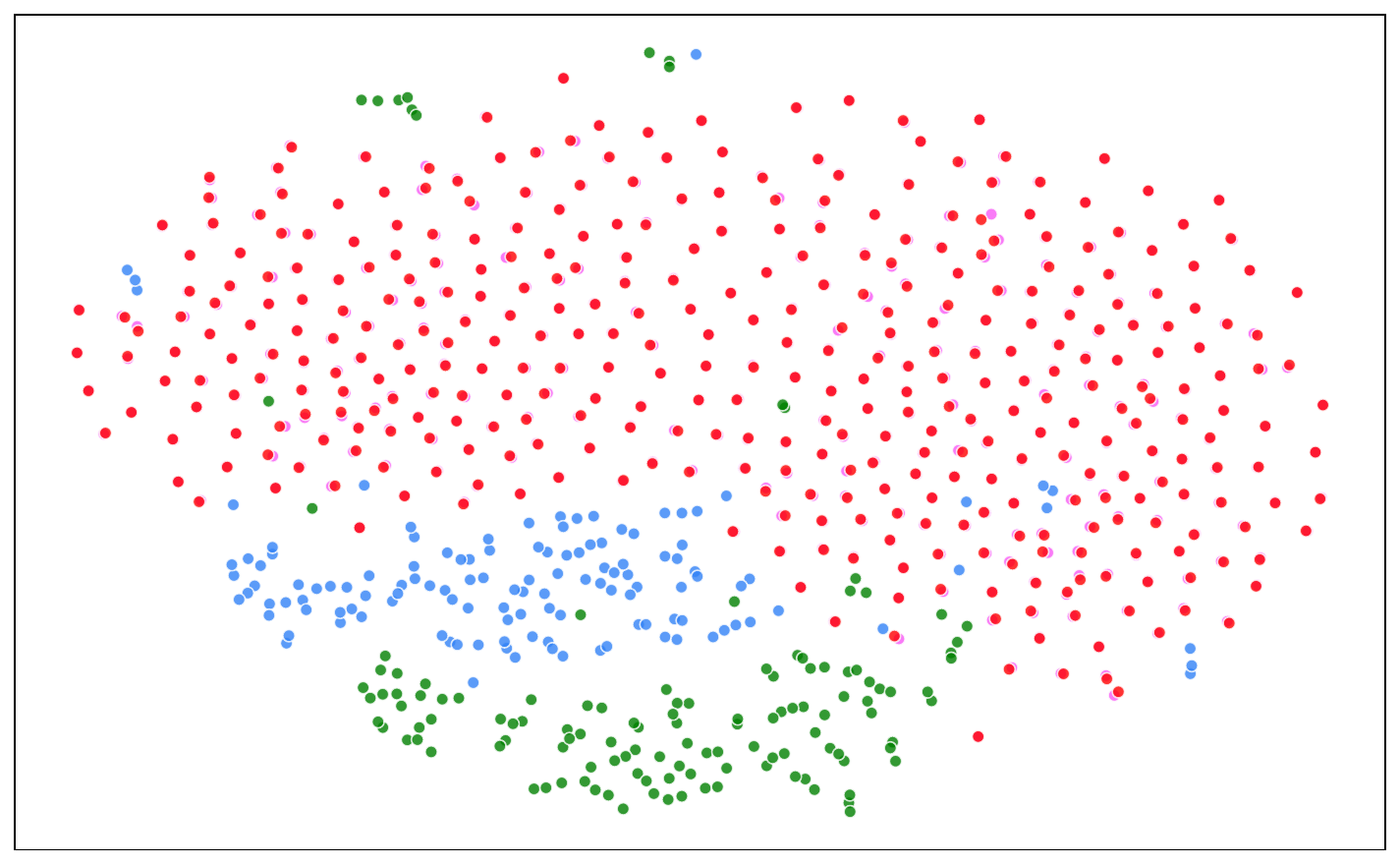}
   \includegraphics[width=0.48\linewidth]{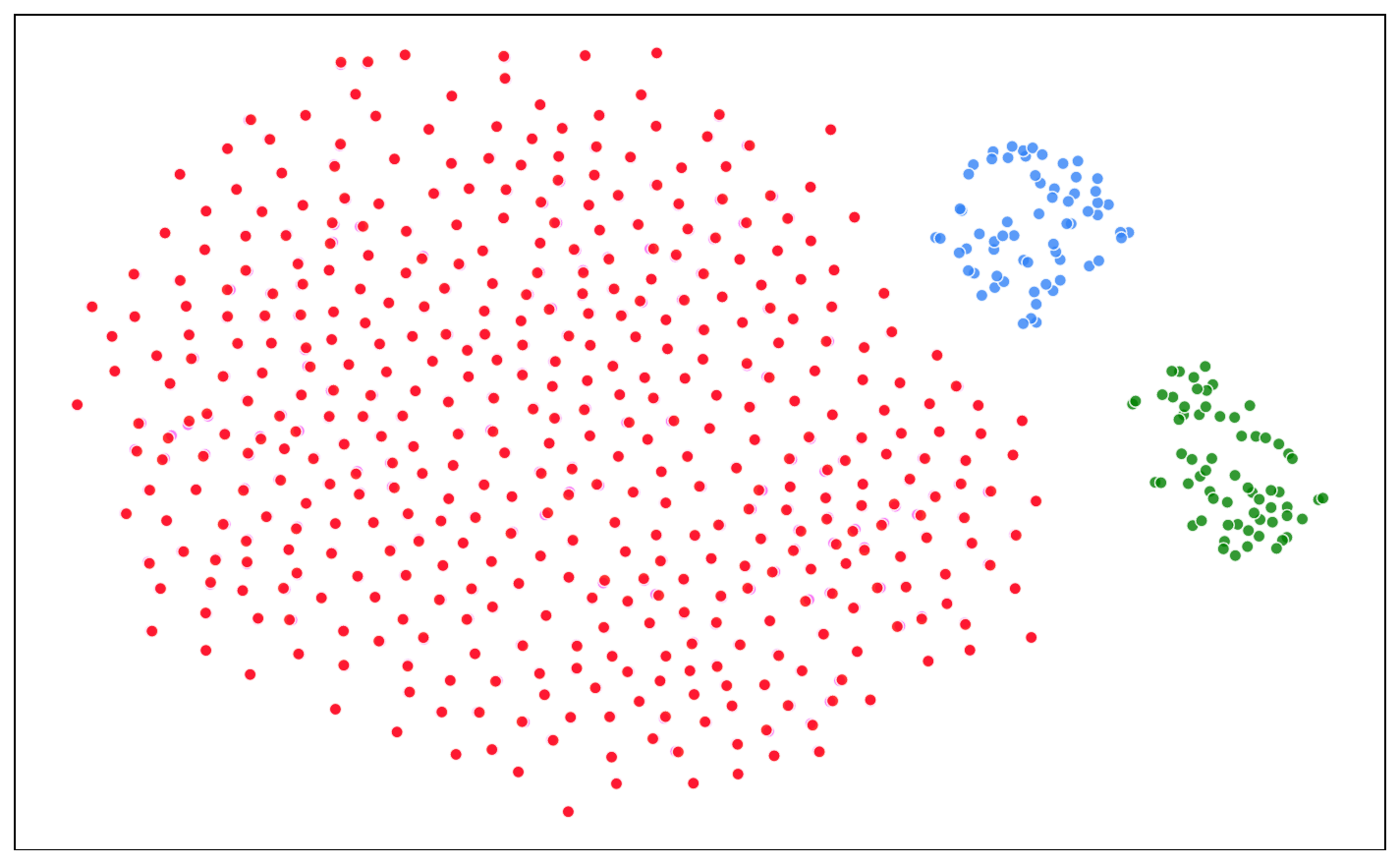}
   \includegraphics[width=0.48\linewidth]{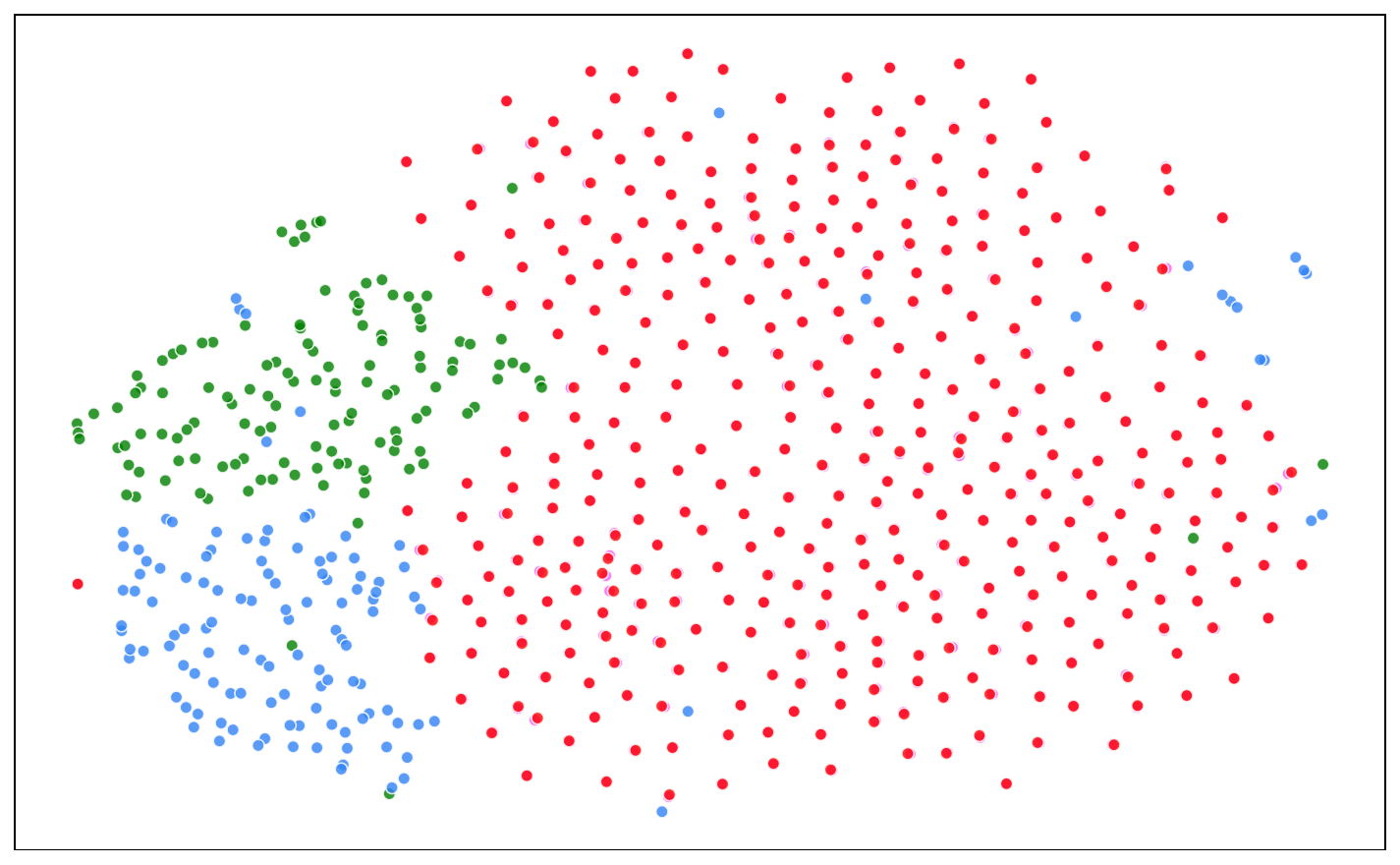}
   \includegraphics[width=0.48\linewidth]{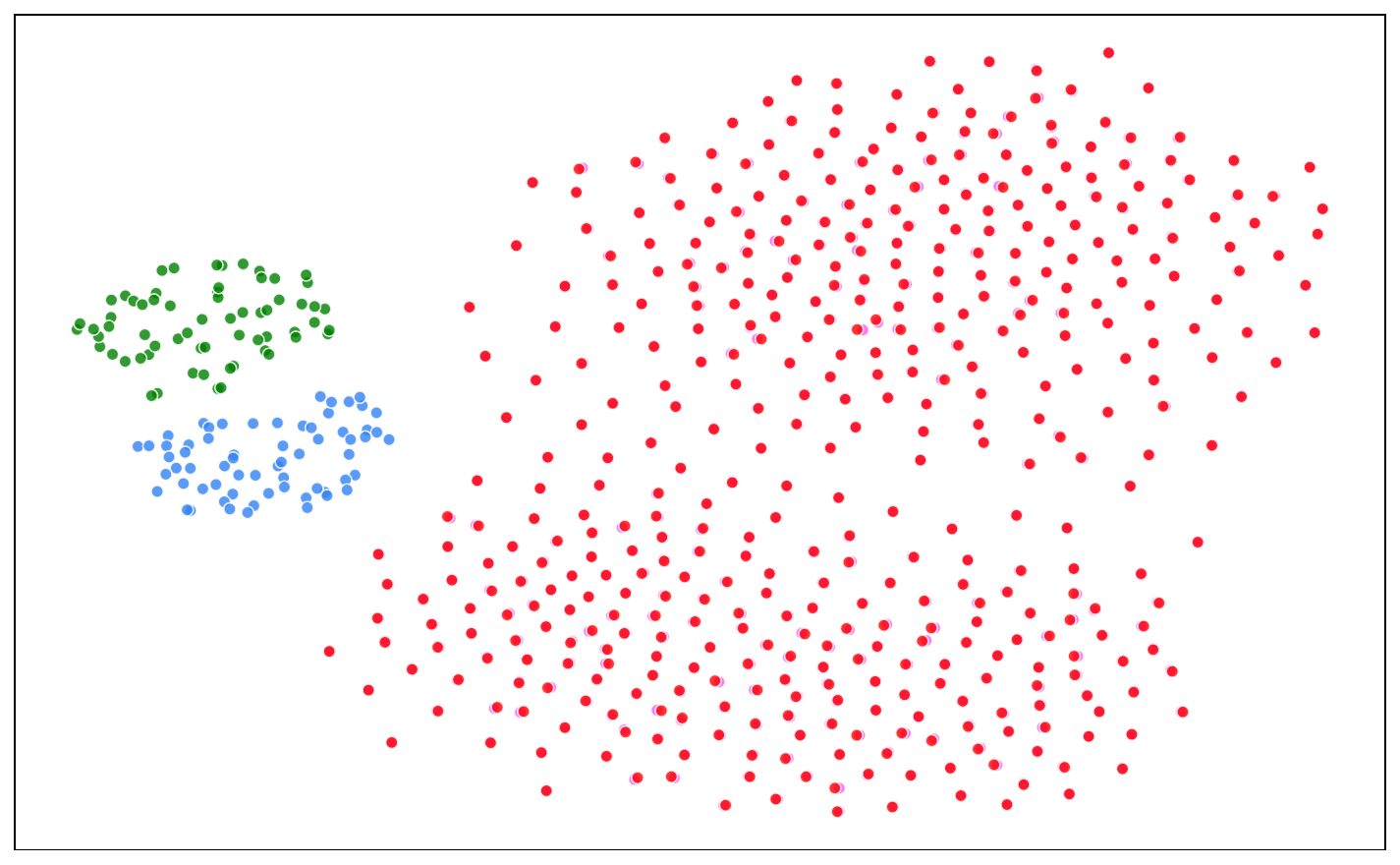}
   \includegraphics[width=0.48\linewidth]{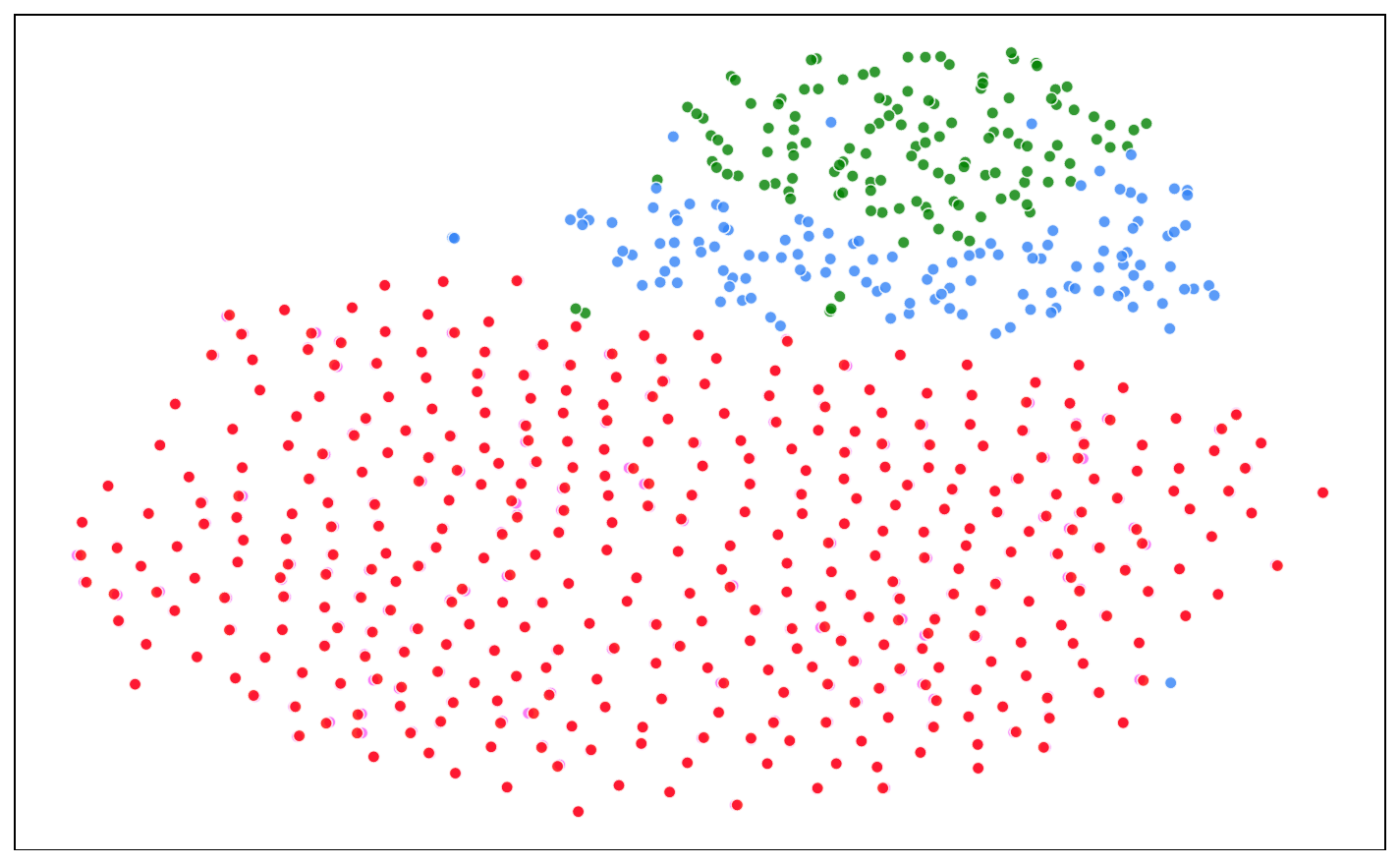}
   \caption{t-SNE representation visualization of 4 batches of samples. Left: SAR-optical; right: RGB-DEM.}
\label{fig:tsne_more}
\end{figure}

See Figure \ref{fig:tsne_more}.

%
%

\end{document}